\pdfoutput=1
\documentclass{article}

\usepackage{amsmath,amsfonts,bm}

\def\eqref#1{equation~\ref{#1}}
\def\1{\bm{1}}

\def\ervx{{\textnormal{x}}}

\DeclareMathAlphabet{\mathsfit}{\encodingdefault}{\sfdefault}{m}{sl}
\SetMathAlphabet{\mathsfit}{bold}{\encodingdefault}{\sfdefault}{bx}{n}

\def\gG{{\mathcal{G}}}

\newcommand{\parents}{Pa} % See usage in notation.tex. Chosen to match Daphne's book.

\usepackage{microtype}
\usepackage{graphicx}
\usepackage{subfig}
\usepackage{booktabs} % for professional tables

\usepackage{hyperref}

\usepackage[accepted]{icml2025}

\usepackage{array}
\usepackage{amsmath}
\usepackage{amssymb}
\usepackage{mathtools}
\usepackage{amsthm}
\usepackage{pdfpages}
\usepackage{wrapfig}
\usepackage{tikz}
\usepackage{adjustbox}
\usepackage{fancyhdr}
\usetikzlibrary{arrows.meta}
\graphicspath{{figures/}}
\usepackage{fancyhdr} % for custom headers and footers

\usepackage{pifont}
\usepackage{xcolor}
\newcommand{\cmark}{\textcolor{green!80!black}{\ding{51}}}
\newcommand{\xmark}{\textcolor{red}{\ding{55}}}

\usepackage[capitalize,noabbrev]{cleveref}

\theoremstyle{plain}
\newtheorem{theorem}{Theorem}[section]

\theoremstyle{definition}
\newtheorem{definition}[theorem]{Definition}

\theoremstyle{remark}

\usepackage[textsize=tiny]{todonotes}

\icmltitlerunning{The CausalMan simulator}

\begin{document}

\twocolumn[
\icmltitle{CausalMan: A physics-based simulator for large-scale causality}




\begin{icmlauthorlist}
\icmlauthor{Nicholas Tagliapietra}{bosch,tud}
\icmlauthor{Juergen Luettin}{bosch}
\icmlauthor{Lavdim Halilaj}{bosch}
\icmlauthor{Moritz Willig}{tud}
\icmlauthor{Tim Pychynski}{bosch}
\icmlauthor{Kristian Kersting}{tud,hessian}
\end{icmlauthorlist}

\icmlaffiliation{tud}{Computer Science Department, TU Darmstadt, Germany}
\icmlaffiliation{bosch}{Bosch Center for Artificial Intelligence, Renningen, Germany}
\icmlaffiliation{hessian}{Hessian Center for AI (hessian.AI), Germany}

\icmlcorrespondingauthor{Nicholas Tagliapietra}{nicholas.tagliapietra@de.bosch.com}

\icmlkeywords{Machine Learning, ICML}

\vskip 0.3in
]



\printAffiliationsAndNotice{\icmlEqualContribution} % otherwise use the standard text.

\begin{abstract}
A comprehensive understanding of causality is critical for navigating and operating within today's complex real-world systems.
The absence of realistic causal models with known data generating processes complicates fair benchmarking. In this paper, we present the \textit{CausalMan} simulator, modeled after a real-world production line. The simulator features a diverse range of linear and non-linear mechanisms and challenging-to-predict behaviors, such as discrete mode changes. 
We demonstrate the inadequacy of many state-of-the-art approaches and analyze the significant differences in their performance and tractability, both in terms of runtime and memory complexity. As a contribution, we will release the \textit{CausalMan} large-scale simulator. We present two derived datasets, and perform an extensive evaluation of both.
\end{abstract}

\section{Introduction}
\begin{figure*}[ht]
\centering
\includegraphics[width=0.95\textwidth]{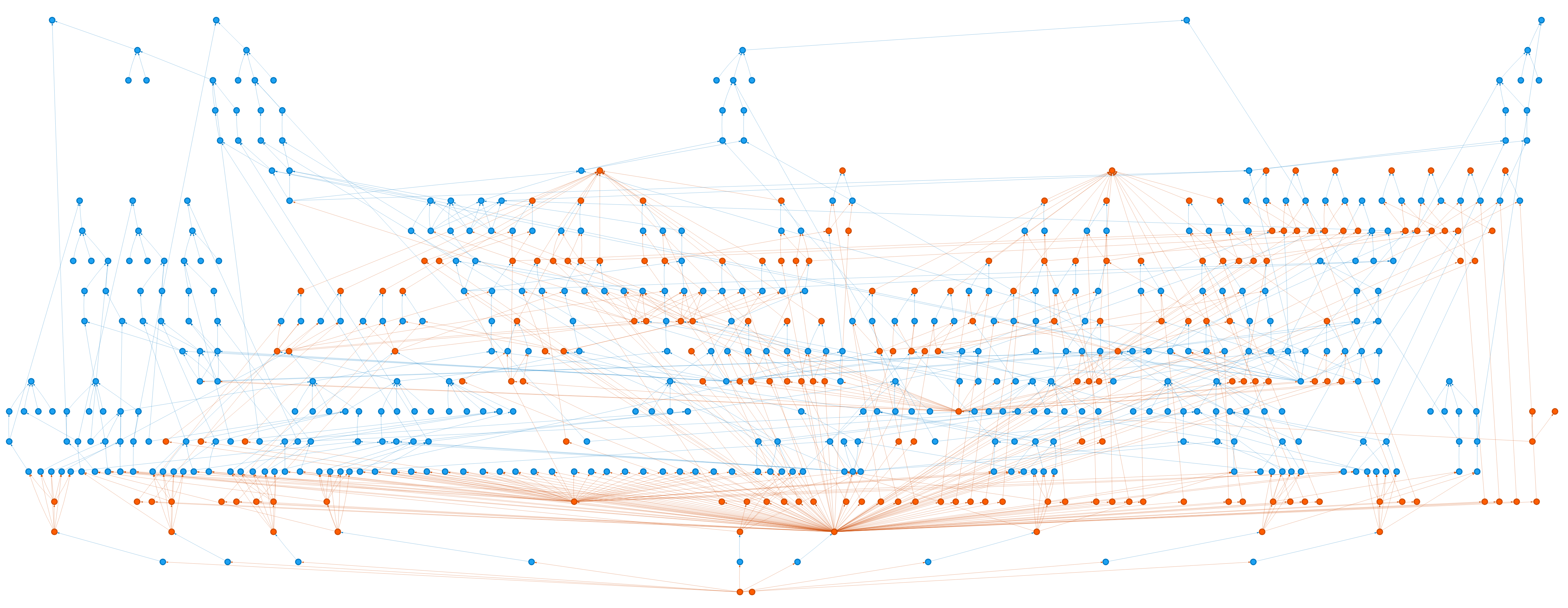}\label{fig:ground_truth_dataset1}
\caption{Complete Ground truth causal graph including hidden variables for CausalMan Medium. Observable variables are colored in orange, and latent ones are colored in blue. 419 of 605 (69.2\%) of variables are latent.}
\end{figure*}
The mastery of \textit{Causal Reasoning} is a long-standing challenge in AI, with the potential to drastically impact many disciplines including medicine, science, engineering, and social sciences. 
The development of agents with an understanding of causality enables them to go beyond statistical co-occurrences, and is connected with desirable abilities such as reasoning and Out-of-Distribution generalization~\citep{richens2024robust}. 
Using the tools of \textit{Causality} \citep{Pearl_2009} we can uncover the \textit{Data Generating Process} (DGP), and manipulate it to gain a better understanding of the system being modeled.
With \textit{Causal Inference} we can estimate the effect of interventions on a system while accounting, among others, for confounding biases and missing data \citep{mohan2019graphicalmodelsprocessingmissing}.
To make progress in this area, a fair and comprehensive evaluation of causal algorithms is crucial, as well as benchmark tests analyzing methods from different angles. 
Laying down a comparison across multiple domains, however, presents various challenges. From a practical perspective, one of the main obstacles that impedes progress in causality is the lack of public benchmarks supporting method evaluation~\citep{Cheng2022EvaluationMA}.
When benchmarking on real world data, the true DGP may be partially or even completely unknown.
Additionally, an individual can either be treated or not, which means that we cannot simultaneously observe both potential outcomes, implying that the ground truth values of the causal estimands are not known. 
Consequently, purely factual observational data is insufficient for evaluation due to the unavailability of counterfactual measurements. 
A similar challenge is indicated by~\cite{DBLP:conf/nips/GentzelGJ19}, who stressed the importance of evaluating on interventional measures and downstream tasks. 
In most cases, however, obtaining interventional data is not possible, unethical, or highly expensive.
Shifting to simulated data, \cite{DBLP:conf/nips/CurthSWS21} argued that algorithms matching the assumptions of the DGP are advantaged in those specific benchmarks, but results may not transfer to other scenarios.
Despite this, when correctly designed, simulation can be a powerful tool to benchmark causal models. 
Thanks to causally-plausible simulators, we can obtain any interventional distribution while retaining control on every parameter knob, with the possibility to study any valuable corner case.
Along this path, we can use simulations to gain insights on the behaviour of causal models at the intersection of non-linearity, causal in-sufficiency and high dimensionality.
For the latter, bringing causality to the large scale has been the main driver for a series of efforts \citep{tigas2022interventions} that tried to understand the scalability issues that several causal models have when dealing with thousands of variables, as well as their inference limitations when performed with finite resources.
Scalability is a challenge not only for inference tasks, but also throughout the whole field of causality. The related task of \textit{Causal Discovery} (CD) i.e.,\ recovering the causal diagram from data, suffers from similar burdens, where often mathematical guarantees are sacrificed in exchange of computational feasibility~\citep{zheng2018dagstearscontinuousoptimization}. 
Hereby, we investigate how those methods perform at large scale, and consequently aim to answer the question whether current approaches are adequate for realistic scenarios. 
Our doubt stems from the looming intractability that current methods possess \textit{by design} \citep{EITER200253} when carrying out certain tasks, both from a theoretical and practical viewpoint. Additionally, differently from other works which explore causality in medicine, genetics and ecology, we focus on the manufacturing domain, which has found only scattered applications in the past \citep{Vukovic2022CausalDI, göbler2024causalassembly}. 
Furthermore, we try to motivate the statement that mathematically sound large-scale causality may require new methodologies and engineering breakthroughs that are not yet developed.

\textbf{Contributions} We begin with a description of the real-world scenario, describing many mechanisms which should not be ignored. Next, we derive a set of mathematical requirements which have to be fulfilled, which motivate the development of the CausalMan simulator, as they are absent in other publicly available datasets.
Specifically, our contribution is three-fold:
\begin{itemize}
    \item We develop a physics-based simulator within the manufacturing domain in close collaboration with domain experts. The simulator is designed to meet the requirements of real-world scenarios and presents challenges such as hybrid data types, causal insufficiency, conditional dependencies, and nonlinearities. We utilize the simulator to generate two large benchmark datasets. We use the simulator to derive two large size datasets. The simulator will be released, enabling researchers to generate new observational and interventional data.
    \item Using the aforementioned datasets, we define various exemplary tasks and observe that a wide range of causal models are computationally prohibitive for certain tasks, while others lack expressiveness.
    \item We perform similar analyses for causal discovery, comparing classic algorithms and recent learning-based methods. 
    \end{itemize}
\section{Related Work}
In this section we analyze related approaches relevant to our work and datasets, highlighting common points and dissimilarities. For more exhaustive surveys on the evaluation of causal models, we address the interested reader to \cite{Cheng2022EvaluationMA}, \cite{DBLP:journals/csur/GuoCLH020} and \cite{DBLP:journals/tkdd/YaoCLLGZ21}.

\textbf{Large-Scale Causality:} In \cite{notallcausalinferencezece}, a theoretical and empirical evaluation on simple causal graphs highlighted the intractability of marginal inference and the scaling laws of different causal models. When the goal is to reduce the complexity of different intractable queries, it is possible to adopt \textit{tractable probabilistic models} such as \textit{Sum-Product Networks} (SPNs) \citep{poon2012sumproductnetworksnewdeep}. Furthermore, it is possible to use SPNs to model causal phenomena \citep{zecevic2021interventional, structuralcausalcircuits, poonia2024chispn, pmlr-v246-busch24a}.

Leveraging its independence from combinatorial objects such as graphs, \textit{Rubin's Potential Outcomes} (PO) framework \citep{Imbens_Rubin_2015} can be used to tackle the scalability problem. However, a notable limitation of the PO framework is its reliance on assumptions like \textit{ignorability}, that is equivalent to unconfoundedness and is not suitable for our strongly confounded use-case.

In the realm of causal discovery, scaling is addressed with novel methodologies such as continuous optimization-based approaches \citep{NEURIPS2018notears, 10.5555/3495724.3497230, DBLP:conf/iclr/LachapelleBDL20} or divide-and-conquer approaches \citep{lopez2022factorgraphs, wu2024sea}. 
However, while easier to scale, they suffer from distinct vulnerabilities. 
~\cite{Reisach2021BewareOT} and ~\cite{Kaiser2021UnsuitabilityON} show that their performance is sensitive to the scale of the data, and can degrade to levels comparable to or worse than classic approaches after data normalization. On a similar note \cite{loh_mse_unsuitable} and \cite{seng2024learning} remarked the limitations of methods relying on mean squared error losses. 
Further, \cite{mamaghan2024evaluationbayesian} studied the drawbacks of common metrics when adopting a Bayesian approach.
%First, results suggest these methods may be exploiting patterns of increasing marginal variance in data to reconstruct a causal graph, without any guaranteed analysis at a Causal level. Consequently, their performance can be controlled by manipulating the scale of the data~\cite{Reisach2021BewareOT, Kaiser2021UnsuitabilityON}, degrading to levels comparable to or worse than classic approaches after data normalization. 
Those drawbacks of ML-Based approaches re-ignited interest in novel and more mathematically grounded methods such as \textit{Extremely Greedy Equivalence Search} (XGES) \cite{nazaret2024extremely} or \textit{Differential Adjacency Test} (DAT) \cite{amin2024scalableflexiblecausaldiscovery}. 
\begingroup
\renewcommand{\arraystretch}{1.25} % Default value: 1
\begin{table*}[t]
\small
\centering
\begin{tabular}{p{2.5cm}|m{1.2cm}|m{1.2cm}|m{1.2cm}|m{1.2cm}|m{1.65cm}|m{1.5cm}} 
 & Nonlinear & Mixed types & Cond. dependencies & Causal insufficiency & Interventional data & Large-scale  \\
\hline
CausalMan (ours) & \cmark & \cmark & \cmark & \cmark & \cmark & \cmark \\
CausalChambers & \cmark & \cmark & \xmark & \xmark & \cmark & \cmark \\
CausalAssembly & \cmark & \cmark & \xmark & \xmark & \cmark & \cmark \\
CausalBench & \cmark & \xmark & \xmark & \cmark & \cmark & \cmark \\
Neuropathic-pain & \cmark & \xmark & \xmark & \xmark & \cmark & \cmark \\
\end{tabular}
\caption{Comparison of CausalMan's main features with other available simulators or datasets.}
\label{table:causalman_comparison}
\end{table*}
\endgroup

\textbf{Datasets and Benchmarks:}
A wide variety of benchmarks for causal models are publicly available~\citep{asiadataset, alarmdataset, sachsdataset}.
However, only a limited number of them target large scale scenarios~\cite{diabetesdataset}, and an even smaller fraction involve hybrid domains, which is the focus of our datasets and experiments.
To compensate the lack of data, a common choice for analysing scaling laws for causal models is to generate random Erdos-Renyi~\citep{Erdos1984OnTE} or Scale-Free graphs~\citep{barabasi1999emergence} which, although easy to simulate, are far from reflecting the real world. 
Recent works provide datasets and methodologies to generate realistic synthetic and semi-synthetic data. 
Semi-synthetic DGPs tuned on real data, often along with the use of prior domain knowledge, are the focus of simulators such as \textit{CausalAssembly} ~\citep{göbler2024causalassembly} for the manufacturing domain, or the \textit{Neuropathic Pain simulator} ~\citep{DBLP:conf/nips/Tu0BK019} in the medical domain. Further, semi-synthetic DGPs are used in \cite{Dorie2017AutomatedVD,Hahn2019AtlanticCI} and \cite{Shimoni2018BenchmarkingFF} to generate datasets with real observational data for the untreated individuals, coupled with simulated treated counterparts.
Contrary to those datasets, our data comprise additional layers of complexity by simulating mechanisms such as batching, hybrid data-types and conditional dependencies.
% Real Data
Concentrating on real world data, CausalBench ~\citep{Chevalley2022CausalBenchAL} is a large scale benchmark for single-cell perturbation experiments with interventional data gathered using gene-editing technologies. 
A different strategy is adopted by CausalChambers~\citep{gamella2024causalchambers}, which builds a real isolated physical system where physical mechanisms are known almost perfectly, giving a high degree of confidence on the exactness of the ground-truth Structural Causal Model. Additionally, \cite{pmlr-v236-mogensen24a, mhalla2020causalmechanismextremeriver} provide real-world datasets with a more or less justified ground-truth causal graph.
\section{Background} 
\label{sec:background}
\subsection{Causal Models} 
Modern causality in the Pearl sense relies on intuitive graphical representations of causal phenomena. 
Here, we assume that the underlying causal structure can be represented by a \textit{Directed Acyclic Graph} (DAG) $\displaystyle \gG = (E,V)$, where the sets $V = \{1, \dots, d\}$ and $E \subseteq V \times V $ are vertices and directed edges respectively. Direct causes of a node $v_i$ are called Parents and are denoted with $\displaystyle \parents_\gG (v_i)$.
We now define \textit{Structural Causal Models}, which incarnate the Pearlian notion of causality \citep{Pearl_2009} and defines the DGP.
\begin{definition} A \textit{Structural Causal Model} (SCM) is a 4-tuple $\mathcal{M} \coloneq (\textbf{U}, \textbf{V}, P_{\textbf{U}}, \mathcal{F})$ where $\textbf{U}$ is the set of exogenous variables that are related to external factors, $\textbf{V}$ is the set of endogenous variables that depend on other endogenous/exogenous ones, $P_\textbf{U}$ is the probability density function of the exogenous variable $\textbf{U}$, and
$\mathcal{F} = \{f_1, f_2, \dots, f_n\}$ is the set of \textit{Structural Equations}, where each element is a mapping such that $f_i : U_i \cup Pa_i \rightarrow V_i$, with $U_i \subseteq \textbf{U}$ and $V_i \subseteq \textbf{V}$. Each endogenous variable is related to a structural equation that determines its values. In practice, each node $v_i \in V$ can be expressed as $ v_i = f_i (u_i, Pa_i) $.
\end{definition}
Interestingly, the dependencies between variables described by the structural equations collectively induce a causal graph. 
%Looking at the dependencies between variables induced by each structural equation, it is possible to extract a causal graph for the phenomena being modeled. 
Additionally, when we assume that the dependency on exogenous variables is additive, i.e.\, in the form $ v_i = f_i (Pa_i) + u_i $, we say that the SCM adopts an \textit{Additive Noise Model} (ANM). 

Causal models can be used to model the effects of \textit{interventions} i.e.\, the manipulation of causal mechanisms. A causal model that is capable of modeling the effect of interventions is called an \textit{interventional model}. Moreover, an \textit{intervention} in a SCM consists in replacing one (or more) structural equation $f$ with a different function $\hat{f}$.When we exchange $\hat{f} = a$ with $a \in \mathbf{R}$, we call it an \textit{atomic} intervention.  An intervention on a SCM might induce a different causal graph than the original unintervened one.
In section~\ref{sec:experiments} we show how different causal models may have radically different properties and computational requirements for the same causal query.

Lastly, even though the complete description of the causal phenomenon is assumed to be a DAG, its marginalisations to lower dimensions may not be DAGs. Indeed, if a set of variables is marked as latent, the operation of marginalizing out latent variables is called \textit{latent projection} \citep{verma2013equivalencecausalmodels}, which can result in a graph containing directed but also bi-directed edges representing relationships without a clear causal direction, called \textit{Acyclic Directed Mixed Graph} (ADMG).
\subsection{Treatment Effect Estimation}
The most common tasks in \emph{Causal Inference} (CI) involves the prediction of the effect of one or multiple interventions on an outcome variable and assess its effectiveness i.e.,\ the \textit{Treatment effect}.
Treatment effect estimation is based on comparing a population of treated individuals with a reference control group that did not receive any treatment. We proceed by defining the \textit{Average Treatment Effect} (ATE) which describes how, on average, an individual responds to a specific treatment:
\begin{equation}
   ATE =  \mathbb{E} [Y(1) - Y(0)],
\end{equation}
where $Y (1)$ and $Y(0)$ indicate respectively the outcomes in presence or absence of a treatment.\\
When searching for fine-grained estimates, we can encounter scenarios where treatments will affect different sub-populations heterogeneously e.g. \textit{Heterogeneous Treatment Estimation} (HTE). To identify the treatment effect to such level of detail, we condition the ATE on $X = x$, and define the \textit{Conditional Average Treatment Effect} (CATE) as
\begin{equation}
    \tau (x) = \mathbb{E} [Y (1) - Y (0) | X = x].
\end{equation}
\section{Motivating scenario} \label{sec:dataset}
Manufacturing lines exhibit complex behaviors that pose significant challenges when performing a causal analysis. For this purpose, we start from a motivating real-world scenario, describing the system and the mechanisms that distinguish an assembly line. Further, we use this scenario to derive a set of requirements that should be fulfilled. Finally, we compare our requirements with publicly available solutions, and discuss where our simulator fits.
\begin{table*}[t]
\centering
\begin{tabular}{ p{1.5cm}|p{1cm}|p{1cm}|p{1cm}|p{1cm}|p{1.5cm}|p{1.5cm}} 
%\hline
 & \multicolumn{2}{|c|}{Full Graph} & \multicolumn{2}{c|}{Observable Graph} & \multicolumn{2}{c}{\#  Samples} \\
\hline
 Dataset & Nodes & Edges & Nodes & Edges & Obs. & Int. \\
\hline
Small & 157 & 121 & 53  & 95(13) & 717,962 & 622,385 \\ 
Medium & 605 & 1,014 & 186  & 381(172) & 717,911 & 620,537\\ 
%\hline
\end{tabular}
\caption{Overview of the two datasets. On the left column we list the information for the full causal graph, while on the right for the partially observable graph. In parentheses we have the number of bi-directed edges. All our experiments use the partially observable (therefore causal insufficient) causal graph.}
\label{table:dataset_recap}
\end{table*}

\subsection{Real-world system}
We study a production line that assembles \textit{magnetic valves} (MV) and \textit{hydraulic blocks} (HB) together, forming \textit{Hydraulic Units} (HU). 

\textbf{Hydraulic Units and Magnetic Valves:} An HU is a device used to control the flow of a fluid. It is composed by an \textit{Hydraulic Block} (HB) and by a certain number of \textit{Magnetic Valves}. An HB is a mechanical component with a different number of bores where, during the assembly process, MVs are inserted into them with a press-fitting machine. 
A \textit{Magnetic Valve} (MV) is the electromechanical component inside the HU thanks to which, after applying a voltage, it is possible to control the flow of a fluid. In practice, by energizing the MVs we can control whether the fluid can flow or not through the HU. Each individual MV and HU could be characterized for their material (elasticity and stiffness) or geometric (length and/or diameter) properties.s), or geometrical quantities (diameter, length).

\textbf{Press-fitting:} The \textit{Press-Fitting} (PF) machine applies a force which inserts the MV into a bore of the HU. The force will insert the MV into the bore, but it will also deform it. At the end of the process the bore will be deeper than before by a certain amount which is determined by the physical models (with some stochasticity). Part of the deformation is permanent, and another other part will disappear after the pressing force is removed at the end of the process, as it is related to the elasticity of the material. If the force is too high, we may cause a damage that will end in the component being scrapped. Faults can be related to the leakage of fluid through the MV and through the HU in situations where it is not supposed to happen. Those faults are often caused by anomalies during the press-fitting process, i.e.\, the pressing force is too high, or can be caused by some properties of the MV or HB not being ideal, making them easier to break during assembly. Further details in \ref{appendix:structural_equations}. 

\textbf{Anomaly detection} \label{sec:dataset_monitoring}
Production lines typically incorporate anomaly detection mechanisms for the purpose of identifying faulty parts that are not fit for use. In the best case scenario, a defective product should be caught soon and removed (scrapped) before reaching the end of the production line. 
Moreover, many attributes have to stay within specific ranges of values (See Fig. \ref{fig:monitoring} in the appendix). This is described with a boolean variable that can be either true or false depending on whether an attribute is within a \textit{Lower Tolerance Limit} (LTL) and an \textit{Upper Tolerance Limit} (UTL). Those tolerances vary depending on the type of component being produced (see Fig. \ref{fig:monitoring} and Sec. \ref{sec:appendix_conditional_depepdences}).
\begin{equation}\text{MpGood}_i = 
\begin{cases} 
True & \text{if } \text{LTL}_i \leq x_i \leq \text{UTL}_i, \\
False & \text{otherwise}.
\end{cases}
\end{equation}
At the end of every process, a \textit{logic AND} operation between every \textit{MpGood} (Mechanic-Part Good) variable is performed to check if all the attributes within the machines fall within the desired range. If that is true, the variable ProcessResult, which signals the quality conformance of the final product, will be \textit{True}, otherwise \textit{False}. If the process result is false, the component is scrapped because at least one of the parameters is not within the acceptable range. 

\textbf{Batching:} Production is subdivided in \textit{batches} i.e.,\ groups of parts being produced together and sharing similar properties. 
On the same production line there might be different batches producing different products. Being a unique production line, all batches share the same causal structure, but they might differ in terms of parametrization. For example, different products may have different geometrical characteristics, or material properties, or a different force applied during press-fitting, etc. %Interestingly, different products identify distinct sub-populations, providing an ideal playground for testing various HTE techniques. %In sec.\ref{appendix:sampling}, we describe how batching affects the sampling procedure for both observational and interventional data. 

\subsection{Key Requirements} \label{sec:dataset_data_requirements}
We translate the expected behavior of a production line into mathematical requirements. Those are:

\textbf{(R1) Large-Scale:} Manufacturing lines have a large number of parameters and sensor measurements which form an intricate causal structure. The majority of those are important and should be considered.
 
\textbf{(R2) Interventional Data:} It should be possible to arbitrarily inject anomalies into the system in the form of an intervention. 

\textbf{(R3) Mixed data-Types:}
Data from manufacturing scenarios includes continuous physical quantities, but also discrete ones such as boolean flags, or categorical identifiers for suppliers and component types. Hence, we need to consider mixed Data-types, e.g.\, continuous, discrete, booleans and categorical variables. 

\textbf{(R4) Conditional dependencies:} Many continuous variables depend on a combination of different discrete parents. For example, the material elasticity of a MV depend on its type and on which supplier produced it, as some suppliers might be better than others.
In mathematical terms, certain node distributions are determined (i.e.,\ caused) by specific combinations of categorical parents. Given a variable $n_i$, the hyper-parameters determining its distribution can change discretely depending on the value of different categorical parent nodes. In Sec.\ref{sec:appendix_conditional_depepdences} a more detailed explanation, and in Fig.\ref{fig:second_order_uncertainty} an illustration of this mechanism. 

\textbf{(R5) Structural Equations and noise models:} Accurate and physically-motivated structural equations enable to model many failure modes that are present in the real system. Most physical models underlying the press-fitting process are nonlinear. This includes the dependency on exogenous variables, hence we are not dealing with any underlying ANM. 

\textbf{(R6) Causal Insufficiency: }\label{sec:dataset_causal_insufficiency}Although all physical mechanisms are well-known, in the real system it is possible to measure only part of the variables, therefore it is necessary to mirror the observability of the real system, marking every simulated variable either as observable or hidden. 

\section{The CausalMan simulator}

We analyse currently available simulators in connection with our requirements.  CausalChambers offers a causal model and realistic data, but is tailored for time-series and does not model large size graphs. The Neuropathic Pain Simulator is both causal and large-scale, and focuses on the medical domain, yet it is limited to binary variables. CausalAssembly is a large-scale simulator with mixed data types which permits the sampling of interventional data. However, it emphasizes causal sufficiency, lacks explicit physical models (which are instead learned), and does not account for conditional dependencies. CausalBench provides a suite of datasets centered on gene expression and interventional data, but it lacks heterogeneous data types, conditional dependencies, and large-scale capabilities. Table \ref{table:causalman_comparison} provides a comparison of those simulators.
Our \textit{CausalMan simulator} has been specifically designed to fulfill the requirements outlined in Sec.\ref{sec:dataset_data_requirements}. CausalMan is based on physical models derived from first principles (described in \ref{appendix:structural_equations}), and includes mechanisms such as conditional dependences (R4) related to suppliers and product types. 
Domain experts have been heavily involved during the entire workflow, including the definition of all physical models (R5) involved in the production life-cycle, and the validation/fine-tuning of simulation hyper-parameters. Suppliers or product types are represented by categorical variables, and physical quantities by continuous ones (R3). Each variable is classified as either observable or latent (R6). Additionally, we simulate a batching mechanism which also influences the sampling process (Sec.\ref{appendix:sampling}). This simulator permits to derive different datasets, reaching possibly hundreds of variables (R1). Since the underlying DGP is based on SCMs, it is possible to arbitrarily apply interventions (R2). For interventions, they can be hard/soft, single, multiple, and even on latent variables.

Lastly, we provide two large-size SCMs obtained from different configuration of the simulator. On \textit{CausalMan Small} we have a DGP of moderate size with 53 variables, whereas on \textit{CausalMan medium} we aim at the large-scale, and provide a DGP with 186 variables. On Table \ref{table:dataset_recap} we provide an overview on the scale of our datasets. 

\section{Experiments} \label{sec:experiments}
In this section we describe the general experimental setting, including the chosen causal models and causal discovery algorithms. Additional implementation details are present in \ref{sec:appendix_implementation}, including how data is pre-processed and numerically embedded.
\subsection{Causal Models}
We perform experiments on a representative set of causal models, with the goal of highlighting the different characteristics that those methods possess by design. We test \textit{Causal Bayesian Networks} (CBN) \citep{bareinboimpch}, \textit{Neural Causal Models} (NCM) \cite{xia2022causalneuralconnectionexpressivenesslearnability}, Normalizing Flows-based models such as \textit{CAREFL} \citep{careflkhemakhem2021} and \textit{Causal Normalizing Flows} \citep{causalnormjavaloy2023}, and \textit{Variational Causal Graph Autoencoders} (VACA) \citep{vacasanchezmartin2021}.
Lastly, when estimating treatment effects, we also consider regression-based techniques such as \textit{Linear} and \textit{Logistic Regression}. 
In appendix \ref{sec:appendix_causal_models} we provide a description of the chosen models. Below, we formulate four different interventional tasks, which reflect possible interventions or anomalies that may occur in the real world. We focus both on the accuracy of the single interventional distribution and also on the final ATE and CATE estimates. Moreover, we remark that all our experiments use the ADMG obtained after a latent projection to marginalize out latent variables (See Table \ref{table:dataset_recap} for more details). 

\textbf{Task 1:} In the \textit{first interventional task}, the treatment is an intervention on a lower tolerance value, which is raised to a higher value, with control value set to a lower one. 
The target variable is discrete, and is a grandparent of the outcome variable, 
\begin{align*}
    \text{ATE} = \mathbb{E}[Y | do(PF\_M1\_T1\_Force\_LTL=18000)] \\- \mathbb{E}[Y | do(PF\_M1\_T1\_Force\_LTL=15000)].
\end{align*}
After the treatment, the true interventional distribution has a higher probability of being 0 compared to the observational distribution, as the window of accepted values is narrower. 
In practical words, the intervention makes all samples to be classified as not good (ProcessResult = False). 

\textbf{Task 2:} We perform a \textit{second interventional task} with the goal of understanding the effect of increasing the press-fitting force (further information in Sec. \ref{appendix:structural_equations}). 
\begin{align*}
    \text{ATE} = \mathbb{E}[Y | do(PF\_M1\_T1\_Force=30000)] \\- \mathbb{E}[Y | do(PF\_M1\_T1\_Force=16000)]
\end{align*}
In this second task, the treatment increases the force value to a very high value, while the control intervention remains in the desired range. For certain types of product, the force is now set outside of the ideal range.
The force variable has multiple bi-directed edges with other variables describing the PF process. Moreover, it is also an ancestor for other variables, therefore an extreme intervention can cause a chain of outlier values to propagate towards other physical quantities that depend on it (For example $s_{grad}$ and $F_{max}$). 

\textbf{Task 3 and 4:} Interventions on parameters may have heterogeneous effects across different sub-populations. Consequently, ATE estimates provide a general insight on the behavior of the system, but cannot capture how different sub-populations react to the treatment, which is why in this case study we adopt a more targeted approach by estimating different CATEs. 
In our dataset, we can think of product types as sub-populations, where interventions on parameters can impact positively the quality of one product while degrading another. 
Therefore, we repeat the same interventional experiments as in task 1 and 2, \textit{while conditioning} on a categorical variable (the product type).

\subsection{Causal Discovery} \label{sec:causal_discovery}
Similarly, we perform Causal Discovery on our datasets using multiple algorithms.
We test classic methods such as the \textit{Peter-Clark} (PC) algorithm \citep{10.7551/mitpress/1754.001.0001}, its variant \textit{PC-Stable} 
\citep{10.5555/2627435.2750365}, and \textit{Linear Non-Gaussian Additive Noise Models} (LiNGAM) \citep{lingam}. For learning-based approaches, we test NOTEARS \citep{DBLP:conf/nips/ZhengARX18}, GOLEM \citep{10.5555/3495724.3497230}, DAG-GNN \cite{pmlr-v97-yu19a} and GranDAG \citep{DBLP:conf/iclr/LachapelleBDL20}. Additionally we capture metrics for a random \textit{Erdos-Renyi DAG} in every experiment to establish how distant those methods are from random guessing. 
\subsection{Metrics}

In a simulated environment, ground truth quantities are available. Therefore, for causal inference tasks we measure the distance between the estimated interventional distributions and the ground truth ones using the Mean Squared Error (MSE), Jensen-Shannon Divergence (JSD) \citep{jensenshannondivergence} and Maximum Mean Discrepancy (MMD) \citep{JMLR:v13:gretton12a}. 
For treatment effects, we measure the MSE between the estimated effect and the ground truth one. 
For causal discovery, we measure the Structural Hamming Distance (SHD), Structural Intervention Distance (SID) \citep{peters2014structuralinterventiondistancesid}, parent-Separation Distance (p-SD) \citep{wahl2024separationbaseddistancemeasurescausal}, Precision and Recall, as described in \ref{sec:appendix_metrics}.
Lastly, we also consider runtime metrics such as training and discovery time, and CPU/GPU memory usage for each model. For reproducibility, each experiment is repeated 5 times with different random seeds. In our results we average each metric across the seeds and report its mean and SD. Additional details on the hardware are listed in appendix \ref{sec:appendix_implementation}.
\section{Results and Discussion}

\begin{table*}[h] 
\small
\centering
\begin{tabular}{ p{1.5cm}|p{1.9cm}|p{1.9cm}|p{1.9cm}|p{1.9cm}|p{1.9cm} }
 %\hline
 Causal Model & ATE MSE & CATE MSE & JS-Div Tr. & MSE & MMD\\
 \hline
 NCM   & 1.115(0.118) & 1.665(0.159) & 0.206(0.005) & \textbf{0.172(0.001)} &  0.259(0.018) \\
 CAREFL & \textbf{0.982(0.223)} & \textbf{1.539(0.635)} & 0.164(0.105) & 0.279(0.197) &  NaN \\
 CNF   & 1.218(0.012) & 1.784(0.082) & 0.297(0.003) & 0.535(0.007) &  NaN \\
 VACA  & 1.214(0.009)   & 1.890(0.163) &\textbf{ 0.163(0.003)} & 0.265(0.006) &   \textbf{0.244(0.009)} \\
 Linear r. & 4.748(0.142)   & - & - & -  &  -\\
 Logistic r.& 0.992(0.015)   & - & - & -  &  -\\
 %\hline
\end{tabular}
\caption{CausalMan Small. Results for the second treatment effect estimation task using $n = 50.000$ samples and the ground truth ADMG. Linear regression is disadvantaged due to the presence of hidden confounders and nonlinear causal mechanisms. Sampling instabilities prevented the evaluation of MMD on CNF and CAREFL.}\label{table:effect_estimation_small_results_3}
\end{table*}

Given the intricate nature of the DGPs involved, it is unknown how most causal models will perform, as the mathematical assumptions on which most causal models rely are not fulfilled. Furthermore, identifiability is likely to not hold anymore. Therefore, we formulate the following research questions: 
\textbf{(Q1)} To what extent do causal models yield accurate estimates in large-scale scenarios? 
\textbf{(Q2)} What is the computational feasibility of implementing causal models in large-scale contexts? 
\textbf{(Q3)} Do available causal discovery methods generate sufficiently accurate causal graphs? 
\textbf{(Q4)} How effectively do causal discovery methods adapt to the challenges presented by large-scale scenarios?
\subsection{Causal Inference} \label{sec:effect_estimation_findings}
\begin{figure}[t]
\centering
\includegraphics[width=0.5\textwidth]{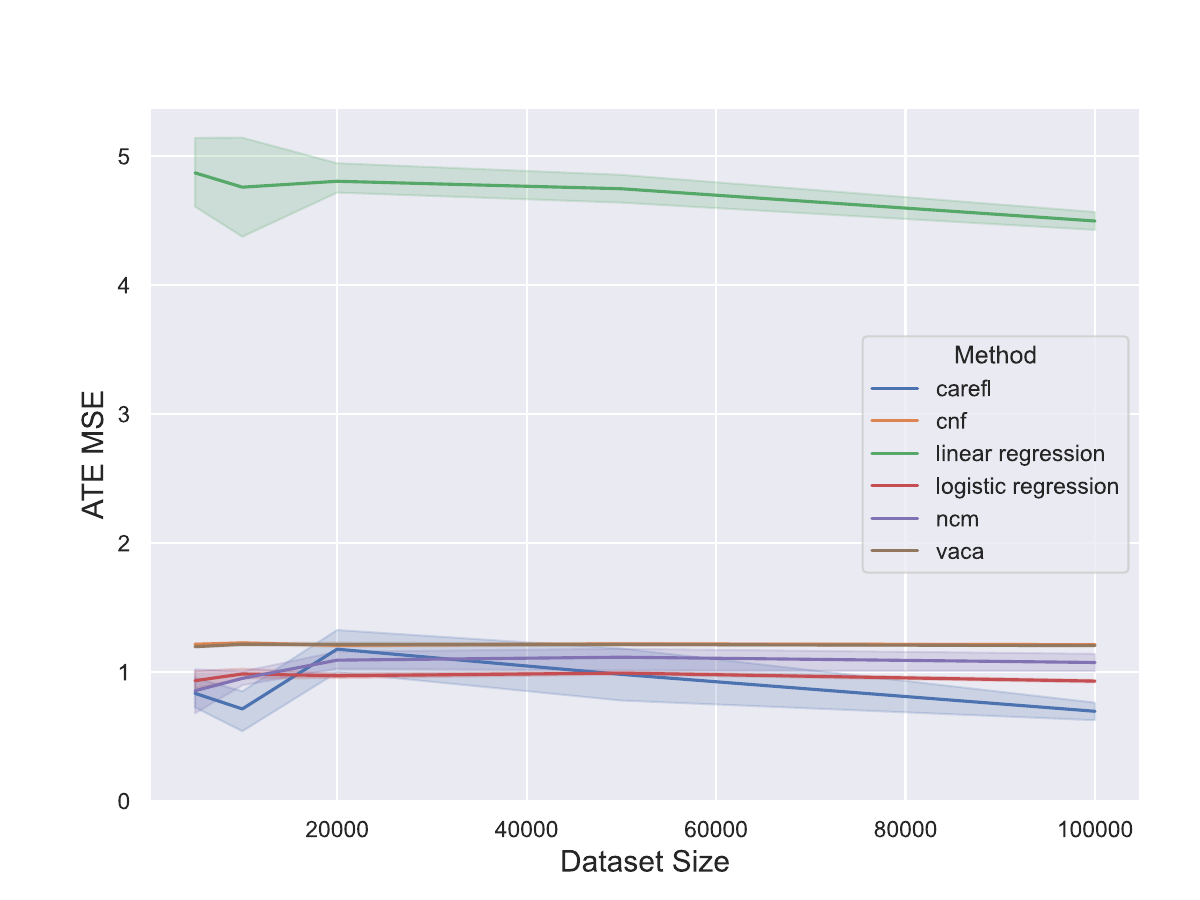} 
\caption{CausalMan Small. Performance for ATE MSE vs. dataset size on the second interventional task. On nontrivial tasks with large amount of nonlinearities and confounders, linear regression is clearly in disadvantage.}
\label{fig:task_3_causalman_small}
\vspace{-0.5cm}
\end{figure}

\textbf{Performance}: Table \ref{table:effect_estimation_results} shows the causal inference performance for the first two effect estimation tasks. 
All models are far from providing a positive answer to our first research question (Q1). 
In fig. \ref{fig:task_3_causalman_small} and table \ref{table:effect_estimation_small_results_3} we can indeed see that all interventional tasks are far from being solved. All causal models fail to reproduce simple interventional distributions. A similar behavior is present when estimating ATEs and CATEs as well.
Furthermore, all results transfer to CausalMan Medium.
On the first interventional task, most models provide inaccurate results, including regression-based methods. On the second task, which deals with an higher amount of confounded and nonlinear causal mechanisms, the deterioration of regression-based methods is evident. Although models based on normalizing flows (CAREFL and Causal Normalizing Flows) are consistently marginally more accurate, they are still unable to provide a satisfactory solution to the new challenges presented by this simulator.
An additional consideration is that, as shown in Figures \ref{fig:ate_mse_vs_size}, \ref{fig:shd_vs_dataset_size_1}, and \ref{fig:shd_vs_dataset_size_2}, all models did not improve significantly with the increase in size of the dataset.

Lastly, in appendix \ref{sec:appendix_additional_task} we study an additional interventional task which, although it is more constructed, studies a corner case where the intervened variable is a direct parent of the outcome. In that scenario, we can observe that a simple linear regression can outperform all other causal models. 

\textbf{Computational Scaling:} From a computational perspective, the results reveal an interesting and diverse landscape of model behavior. Methods based on normalizing flows or regression can provide a positive answer to (Q2), while all other learning-based models require prohibitive amounts of compute.
The computationally heaviest models are CBNs. For CBNs, which are capable of handling only discrete variables, continuous variables have been uniformly quantized in a finite number of steps. However, this design choice is associated with an explosion in memory requirements during the fitting process. This is due to the combination of a high number of states and the in-degree (e.g,\ parents) of some nodes, which leads to an exponential increase in the number of conditional probability distributions to be estimated. To limit memory requirements, we restrict the number of quantization steps to 20, as a higher number would lead to memory demands that are impossible to satisfy. 
No experiments were possible on CausalMan medium for the same reason, even after aggressively quantizing the training data.

Contrarily, deep models follow different scaling laws, as their complexity is mainly related to their architecture and the number of parameters in the network, rather than to the number of nodes. % in the network. 
In other words, large-scale causality does not directly imply a higher number of parameters, but larger causal graphs may require a higher model capacity, and consequently bigger neural networks. 
Among deep models, NCMs are proven to be the most computationally expensive. 
Figure~\ref{fig:runtime_vs_dataset_size} shows a long runtime and significant memory demands (Fig. \ref{fig:memory_usage_dataset1}) for training, thus limiting possible applications to large-size causal graphs. CAREFL and CNF showed more convenient scaling laws with respect to the dataset size (Fig. ~\ref{fig:runtime_vs_dataset_size}), and lower memory requirements (Fig. \ref{fig:memory_usage_dataset1}) with CausalMan medium. Overall, CAREFL and CNF have the potential to scale (computationally) to an even higher number of nodes, whereas NCM appear limited.  

\subsection{Causal Discovery} \label{sec:findings_causal_discovery}
\begin{figure}[t]
\includegraphics[width=0.5\textwidth]{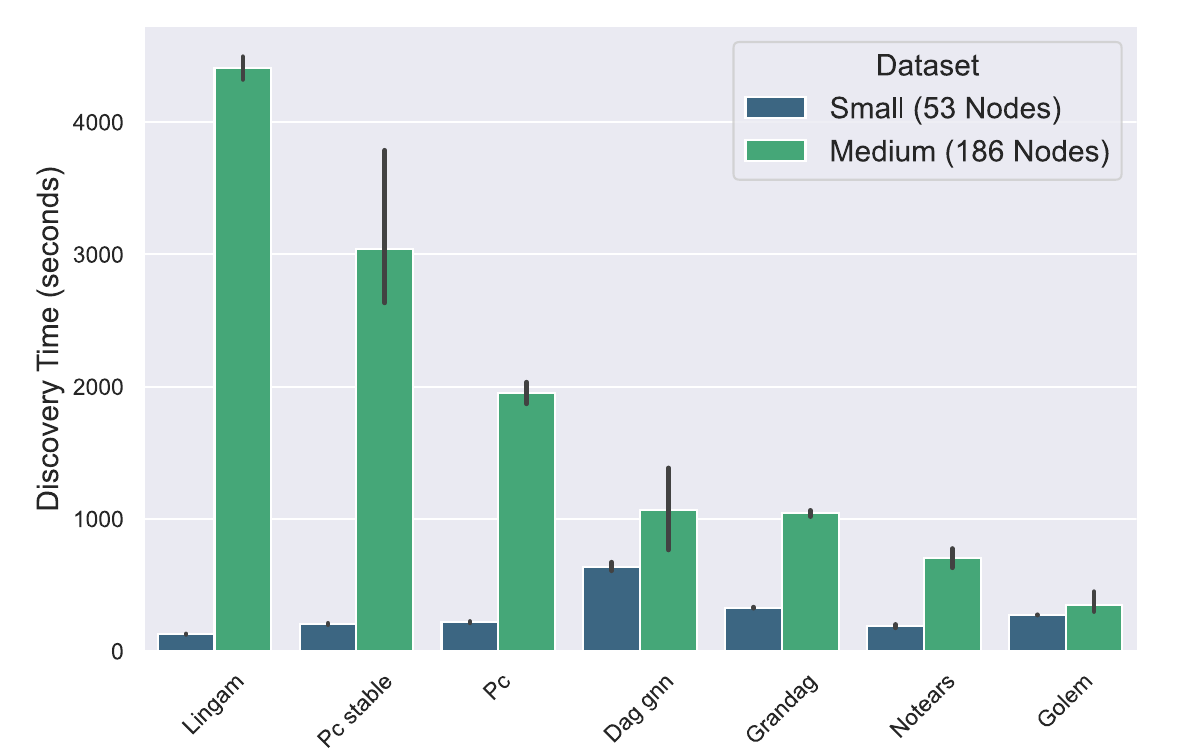}
\caption{Time to discover a Causal graph with $n = 10.000$ samples. Methods thriving on CausalMan Small may be computationally impractical on CausalMan Medium.}
\vspace{-0.5cm}
\label{fig:discovery_time_10000_double_bar}
\end{figure}

Tables \ref{table:causal_discovery_small} and \ref{table:causal_discovery_medium} show results for causal discovery, and Sec.~\ref{sec:appendix_additional_results} provides additional results. 
Similarly, the answer to (Q3) is also negative. All algorithms are far from providing an accurate reconstruction of the causal graph on both datasets. 
Interestingly, we can see in tables \ref{table:causal_discovery_small} and Fig. \ref{fig:shd_bar_comparison}, that CausalMan Small constitutes an intermediate ground where classic methods such as PC or LiNGAM algorithms remain competitive with ML-Based methods. In this dataset, classic methods can still manage the dimensionality of the problem, both performance-wise and resource-wise.  In contrast, when scaling to CausalMan Medium, their limitations are visible in Fig. \ref{fig:shd_bar_comparison} and \ref{fig:discovery_time_10000_double_bar}, where classic methods fall behind.
Moreover, the SHD performance of all methods on all datasets is almost independent of the dataset size (Figure  \ref{fig:shd_vs_dataset_size_1} and  \ref{fig:shd_vs_dataset_size_2}), suggesting a limited capacity of leveraging large amounts of data.

Resource-wise, the answer to (Q4) is negative for classic methods, while learning-based methods might provide a viable approach. We observe strong scaling issues for constraint-based methods in Fig.\ref{fig:discovery_time_10000_double_bar}, where their runtime is multiplied by 20 to 40 times upon tripling the nodes in the graph. The decreasing performance of the PC algorithm on the second dataset can also be explained by the inapplicability of conditional independence tests on large graphs, as the probability of finding a d-separating set is infinitesimal as the number of variables tends to infinity~\citep{feigenbaum2023unlikelihooddseparation}. For methods based on continuous optimization, the growth in compute requirements is significantly smaller, and scaled almost linearly with the number of nodes.
Additionally, Fig.~\ref{fig:runtime_vs_dataset_size} indicates additional scaling limitations, this time with respect to the dataset size, where another significant increase in computation resources occurs. As before, this is not a problem for learning-based methods.
Our analysis demonstrates that current CD methods, when dealing with large graphs, can only be part of an exploratory analysis, and are still far from providing a stand-alone method for reconstructing an accurate causal diagram. Moreover, our results support that the current best approach relies on an iterative \textit{human-in-the-loop} process, based on the combination of CD methods and expert knowledge. 
\section{Conclusions}
We introduced a simulator in the manufacturing field, from which two novel datasets are extracted. The simulator is based on physical models derived from first principles, and the integration of domain knowledge from experts received the highest priority when building the DGP. We envision that our benchmarks will serve as a playground to build causal models that can tackle the complexity of the real world, where most assumptions made by causal models no longer hold. Moreover, although much progress has been made in causal modeling, our research questions received mostly negative answers. 

\textbf{Limitations:} From a simulation perspective, although we modeled the system with a high degree of realism, it still inherits all the modeling assumptions of the underlying SCMs. From a benchmarking perspective, since the models performed far from optimal, we did not test the most complex queries possible, as they are out of reach for all tested models
Furthermore, accurate estimates of ATE or CATE may not always be enough to satisfy real-world use cases. 

\newpage
\section*{Impact Statement}

Releasing simulators and datasets with a high degree of realism has often limitations due to privacy and confidential reasons. Indeed, especially in the industry, it can be challenging to make such data available without incurring in a leak of important information. For those reasons, there are only a limited number of public datasets which can offer a comparable degree of realism, and peculiar mechanisms.  
Consequently, releasing this simulator has the potential to stimulate research in large-scale causality. Moreover, the widespread implementation of causal models in real-world scenarios can yield significant societal benefits. Causal models, compared to non-causal ones, provide more fairness and interpretability, enhance out-of-distribution generalization, and ultimately contribute to the development of safer and more robust systems. 

\section*{Acknowledgments}

We acknowledge support from Robert Bosch GmbH and from the HMWK project “The Third Wave of Artificial Intelligence - 3AI”.

\bibliography{icml2025}
\bibliographystyle{icml2025}

%%%%%%%%%%%%%%%%%%%%%%%%%%%%%%%%%%%%%%%%%%%%%%%%%%%%%%%%%%%%%%%%%%%%%%%%%%%%%%%
%%%%%%%%%%%%%%%%%%%%%%%%%%%%%%%%%%%%%%%%%%%%%%%%%%%%%%%%%%%%%%%%%%%%%%%%%%%%%%%
% APPENDIX
%%%%%%%%%%%%%%%%%%%%%%%%%%%%%%%%%%%%%%%%%%%%%%%%%%%%%%%%%%%%%%%%%%%%%%%%%%%%%%%
%%%%%%%%%%%%%%%%%%%%%%%%%%%%%%%%%%%%%%%%%%%%%%%%%%%%%%%%%%%%%%%%%%%%%%%%%%%%%%%
\newpage
\appendix
\onecolumn

\section{Datasets Release}
All the data used in this paper, and more, is available at this link: \href{https://zenodo.org/records/13871097}{Link to Zenodo anonymous repository.}

\section{Additional task} \label{sec:appendix_additional_task}
In this section we study an additional task, which is more constructed than the others, and is analyzes a corner case where results appear drastically different than in the other tasks.
\begin{equation}
    ATE = \mathbb{E}[Y | do(PF\_M1\_T1\_Force\_MpGood=0)] - \mathbb{E}[Y | do(PF\_M1\_T1\_Force\_MpGood=1)]
\end{equation}
On this task, we are intervening directly on the outcome's direct parent. The outcome variable has only binary parents, and its structural equation is computing the logic AND between all of them. The treatment is setting one of those parents to 0, which leads to the outcome distribution to be 0 with 100\% probability. The control group instead is set to 1. As before, the outcome variable is $Y = Sec\_C2\_Machine1\_ProcessResult$. When conditioning, the evidence variable is still $HU\_HU\_Block\_Type\_ID\_num$, which will be assumed to be observed with value $921$.

\begin{figure*}[t]
\centering
\subfloat[]{\includegraphics[width=0.48\textwidth]{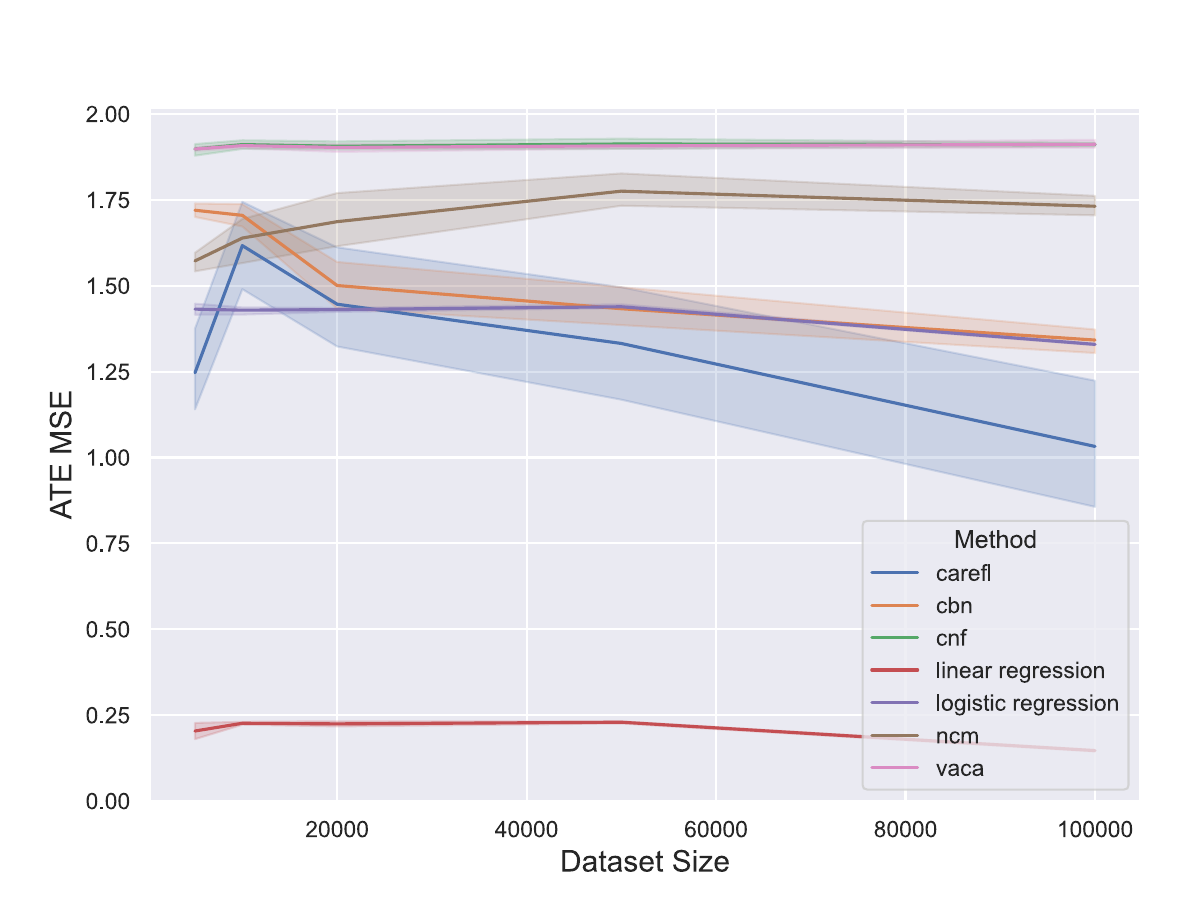} \label{fig:ate_mse_vs_size}}
\subfloat[]{\includegraphics[width=0.52\textwidth]{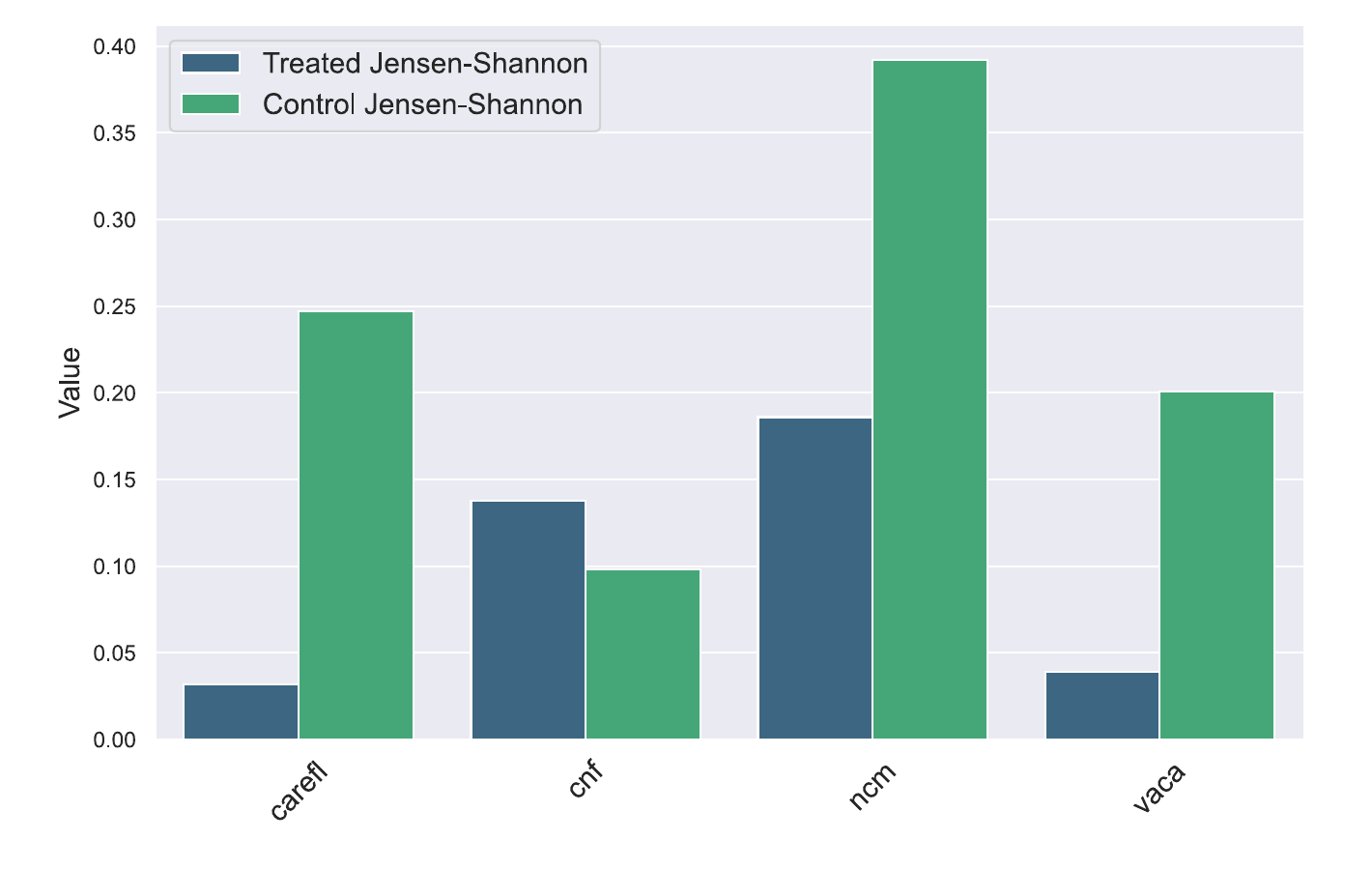}\label{fig:jensen_shannon_comparison}}
\caption{CausalMan Small. Figure \ref{fig:ate_mse_vs_size} shows a stagnation in performance for effect estimation, even with the use of more data. 
Figure \ref{fig:jensen_shannon_comparison}, instead, illustrates the JS-Div. accuracy of treated and control distributions for learning-based causal models, after training with $n = 50.000$ samples.}
\end{figure*}

\begin{figure*}[ht]
\centering
\includegraphics[width=0.8\textwidth]{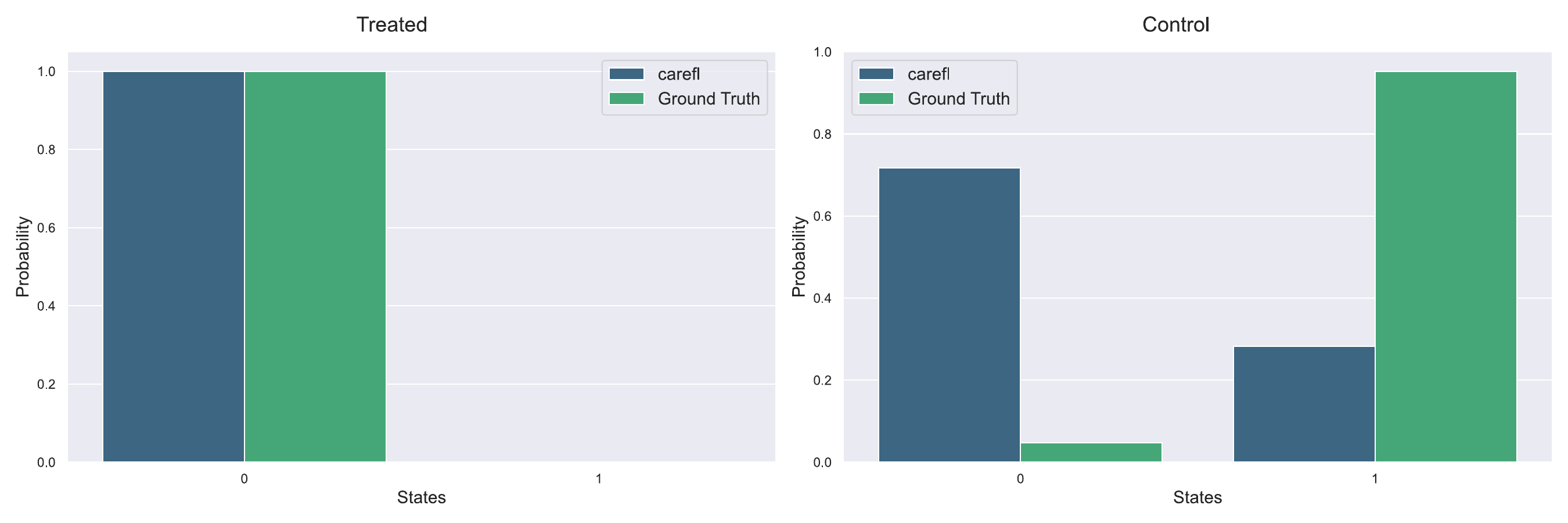}
\includegraphics[width=0.8\textwidth]{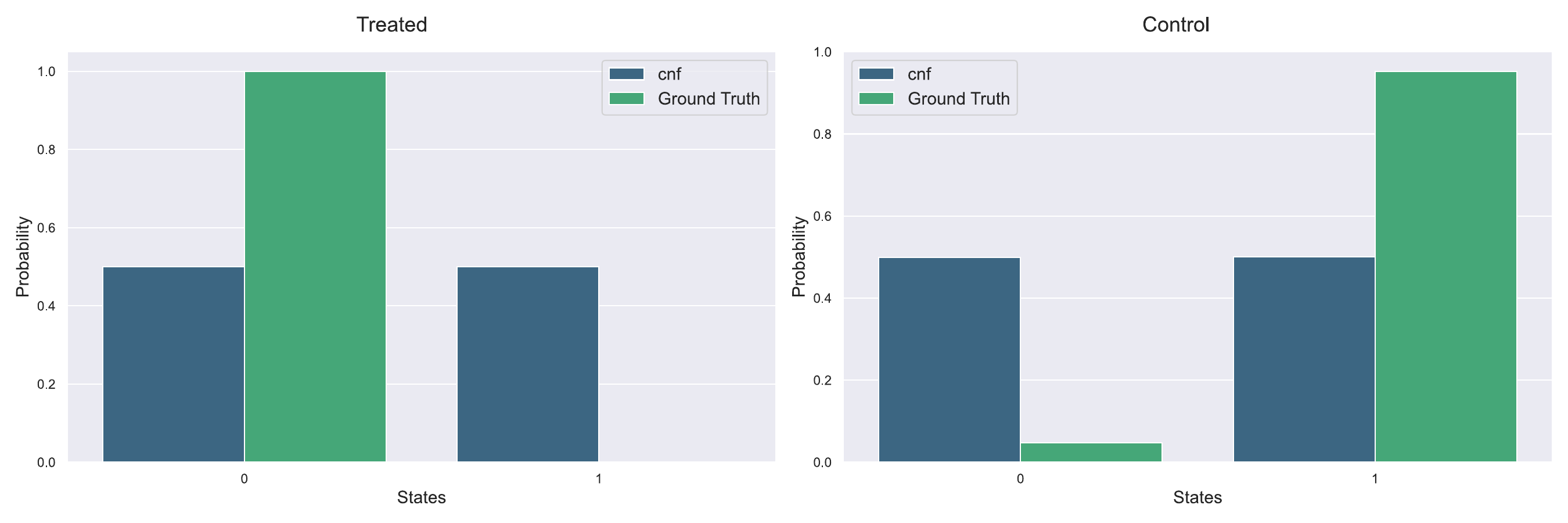}
\includegraphics[width=0.8\textwidth]{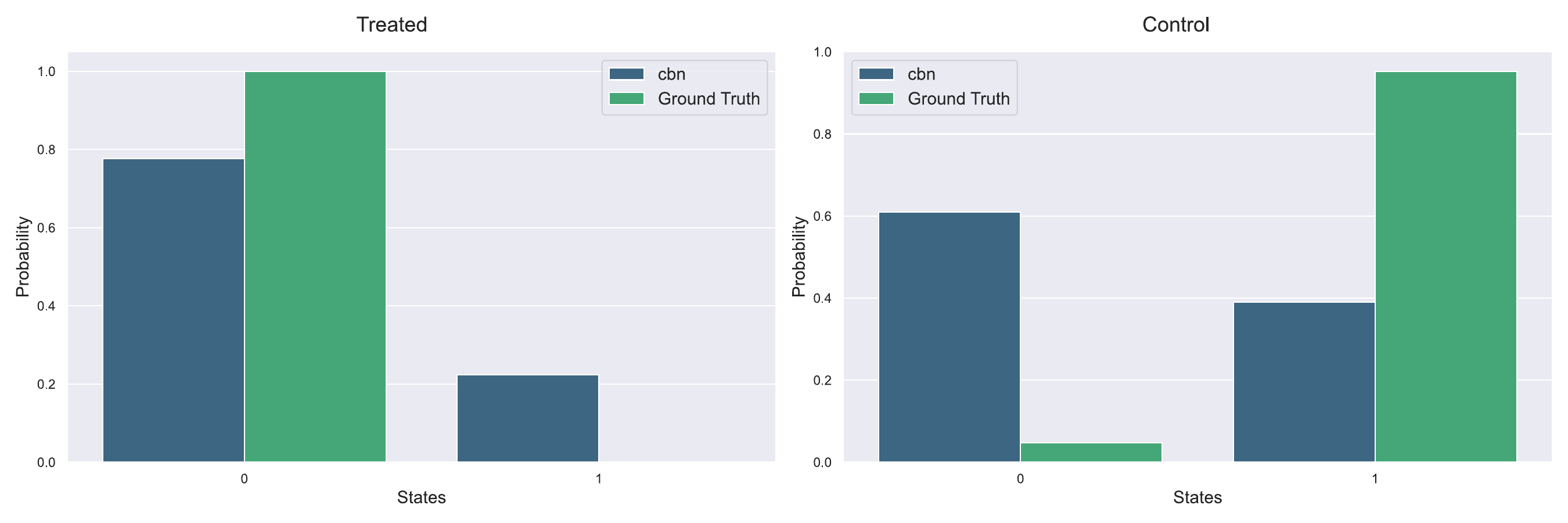}
\caption{Estimated Interventional distributions with CAREFL, CNF and CBN for the additional interventional task (described in \ref{sec:appendix_additional_task}) on CausalMan Small (20.000 samples, seed 42). Causal models are not consistent when estimating interventional distributions, and cannot provide accurate reconstructions of both treated and control populations at the same time.}
\label{fig:interventional_distributions_1_task_1_causalman_small}
\end{figure*}

\begin{figure*}[ht]
\centering
\includegraphics[width=0.8\textwidth]{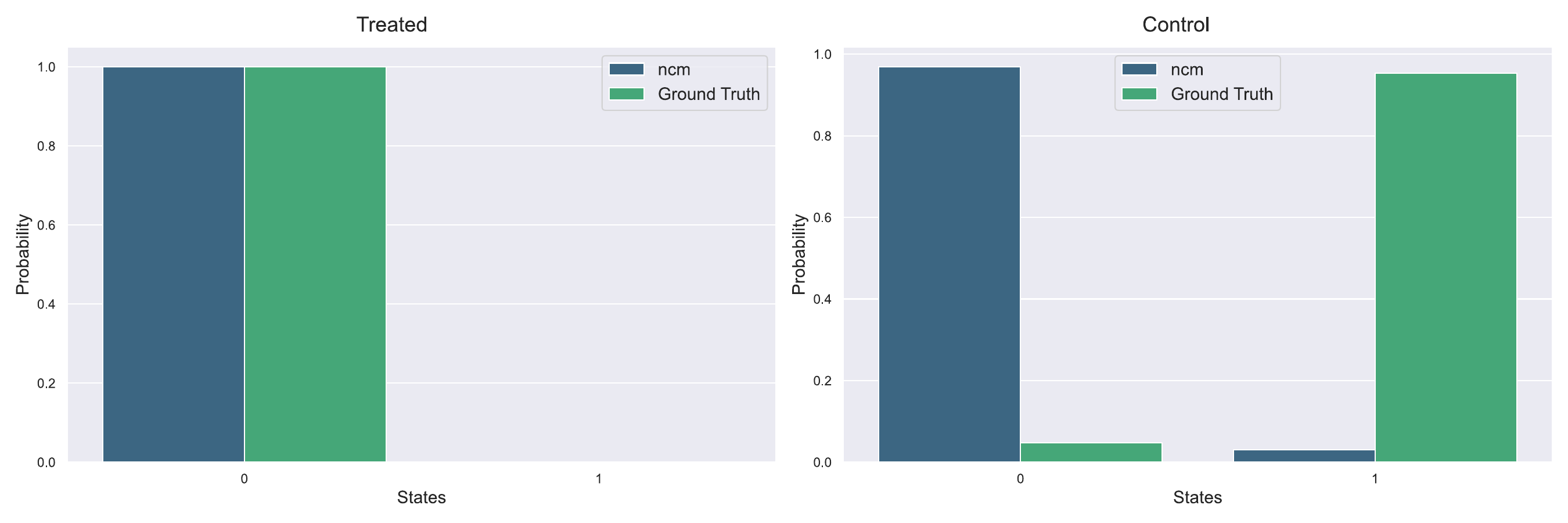}
\includegraphics[width=0.8\textwidth]{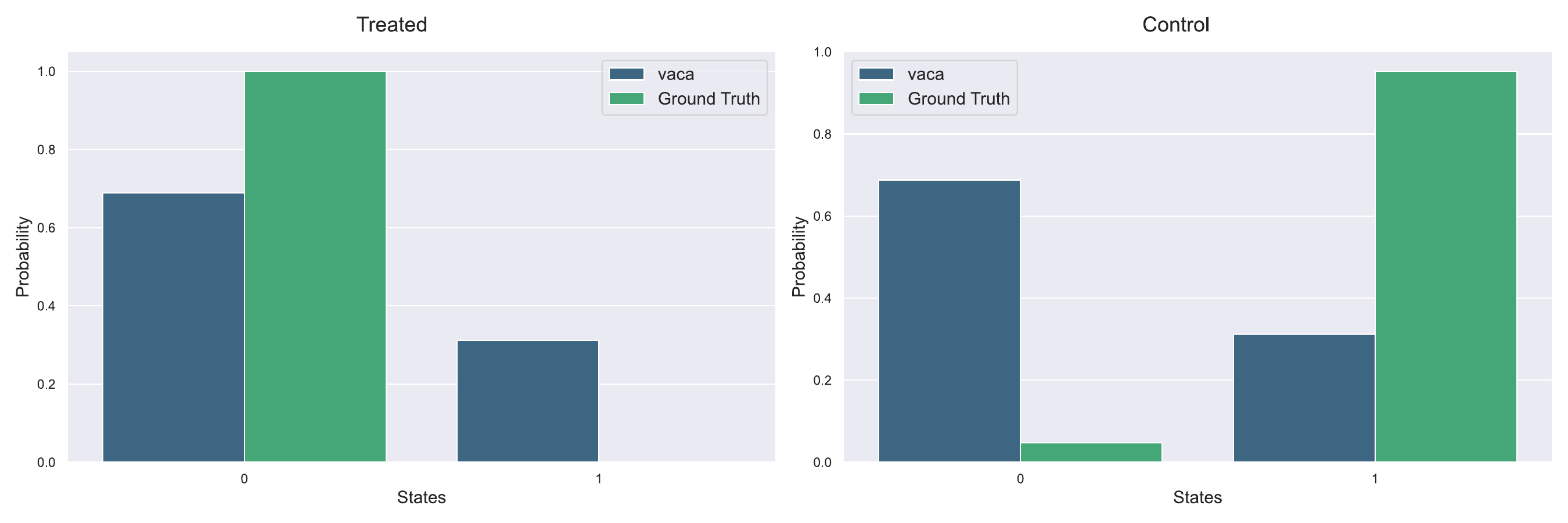}
\caption{Estimated Interventional distributions with NCM and VACA for the additional interventional task (described in \ref{sec:appendix_additional_task}) on CausalMan Small (20.000 samples, seed 42). Causal models are not consistent when estimating interventional distributions, and cannot provide accurate reconstructions of both treated and control populations at the same time.}
\label{fig:interventional_distributions_2_task_1_causalman_small}
\end{figure*}

Surprisingly, we observe in Fig.\ref{fig:ate_mse_vs_size} and table \ref{table:effect_estimation_results} how a simple linear regression outperforms all other causal models. To understand this result, it is essential to notice that the intervened variable is on the \textit{markov blanket} of the outcome, making this behavior expected in a SCM-based DGP. Moreover, we notice that for every causal model, apart from regression-based techniques, ATE or CATE is not estimated directly. We are estimating the empirical treatment effect.
Indeed, in those models, treatment effects are estimated by averaging over samples from the interventional distributions for treated and control populations. Interestingly, deep causal models exhibit superior performance when estimating the treated interventional distributions while being highly inaccurate for treatment effect estimation (Fig.\ref{fig:interventional_distributions_1_task_1_causalman_small} and \ref{fig:interventional_distributions_2_task_1_causalman_small}). 
This can be explained by looking at the discrepancy between the JS-Divergence of the reconstructed interventional distributions for the treated and control groups. In Fig.\ref{fig:interventional_distributions_1_task_1_causalman_small}, \ref{fig:interventional_distributions_2_task_1_causalman_small} and \ref{fig:jensen_shannon_comparison} it can be seen that, even though the treated population is often accurate, the control population is instead mostly poorly estimated. However, accurate treatment effect estimation using those models require precise reconstructions of both treated and control distributions, and the best-performing models overall are simple regression-based techniques that do not go through this procedure and target ATE or CATE directly.

\begin{table*}[t] 
\centering
\begin{tabular}{ p{1.5cm}|p{1.9cm}|p{1.9cm}|p{1.9cm}|p{1.9cm}|p{1.9cm} }
 %\hline
 Model & ATE MSE & CATE MSE & JS-Div Tr. &  MSE &  MMD\\
 \hline
 CBN   & 1.433(0.061)  & 1.653(0.035) & 0.319(0.002) & 0.742 (0.003) &  0.734(0.116) \\
 NCM   & 1.75(0.068)   & 1.502(0.141) & 0.589 (0.000) & 1.000 (0.000) &  0.396(9.023) \\
 CAREFL & 1.332 (0.211)   & \textbf{1.574(0.288)} & 0.512 (0.093) & 0.939 (0.088) &  \textbf{0.035 (0.087)} \\
 CNF   & 1.913(0.018)   & 1.8(0.04) & \textbf{0.291(6e-5)} & 0.707 (0.000)&   Nan \\
 VACA  & 1.907(0.009)   & 1.974(0.274) & 0.332(0.01) & \textbf{0.339 (0.005)}&  0.319(0.009) \\
 Linear r. & \textbf{0.229(0.004)}   & - & - & - & -  \\
 Logistic r.& 1.439(0.008)   & - & - & - &  - \\
 %\hline
\end{tabular}
\caption{Comparison for the additional task on CausalMan Small with $n = 50.000$ samples and ground truth ADMG. Instabilities during sampling prevented to evaluate MMD for CNF, as multiple datapoints diverged to $+\infty$.}.
\label{table:effect_estimation_results}
\end{table*}
\section{Sampling}\label{appendix:sampling}
The batching mechanism affects how data is sampled. First, the production line is unique, and all batches share the same causal graph. What varies between batches is their parametrization. For example one batch may be related to product A, while another batch may be producing product B. Since product A has certain characteristics, the underlying SCM will have a specific parametrization related to product A. Similarly, samples related to product B will be sampled from an SCM with a parametrization related to product B. 

Therefore, batches have to be sampled separately. Within a batch we can perform \textit{ancestral sampling} \citep{koller} on the SCM \textit{related to the batch}. 
In practice, for every batch we set one parametrization of the SCM, and only then we perform ancestral sampling. For next batches we repeat this procedure by setting new parameters on the SCM and then sampling again. This is one of the mechanisms that give rise to the discrete mode changes described in \ref{sec:dataset_data_requirements}.

\textbf{Interventional data:} Interventions are defined within a batch, and Interventional data is sampled by first setting the correct SCM parameterization relative to the batch, and then applying the hard/soft intervention. 
Next, ancestral sampling  is performed as for observational data. 
In other words, we have \textit{Interventional Batches} where a batch is sampled while an intervention is being applied. This procedure is also applied when sampling the ground truth data for treated and control groups during the treatment effect estimation experiments.

\section{Causal Mechanisms} \label{sec:appendix_conditional_depepdences}

We proceed by describing the details of the DGP.

\subsection{Conditional Dependencies:} 
\begin{figure}{
\centering
\adjustbox{valign = c}{\begin{tikzpicture}
    % Define nodes
    \node (A1) at (0, 2) {$A = a_1$};
    \node (B1) at (3, 2) {\textcolor{blue}{$B|A=a_1 \sim \mathcal{N}(\mu_1, \sigma_1^2)$}};

    \node (A2) at (0, 1) {$A = a_2$};
    \node (B2) at (3, 1) {\textcolor{red}{$B|A=a_2 \sim \mathcal{N}(\mu_2, \sigma_2^2)$}};

    \node (A3) at (0, 0) {$A = a_3$};
    \node (B3) at (3, 0) {\textcolor{green}{$B|A=a_3 \sim \mathcal{N}(\mu_3, \sigma_3^2)$}};
    
    % Draw arrows
    \draw[->, double distance=2pt, -{Latex[length=2mm, width=3mm]}] (A1) -- (B1);
    \draw[->, double distance=2pt, -{Latex[length=2mm, width=3mm]}] (A2) -- (B2);
    \draw[->, double distance=2pt, -{Latex[length=2mm, width=3mm]}] (A3) -- (B3);
\end{tikzpicture}}}
\adjustbox{valign = c}{\includegraphics[width=0.55\textwidth]{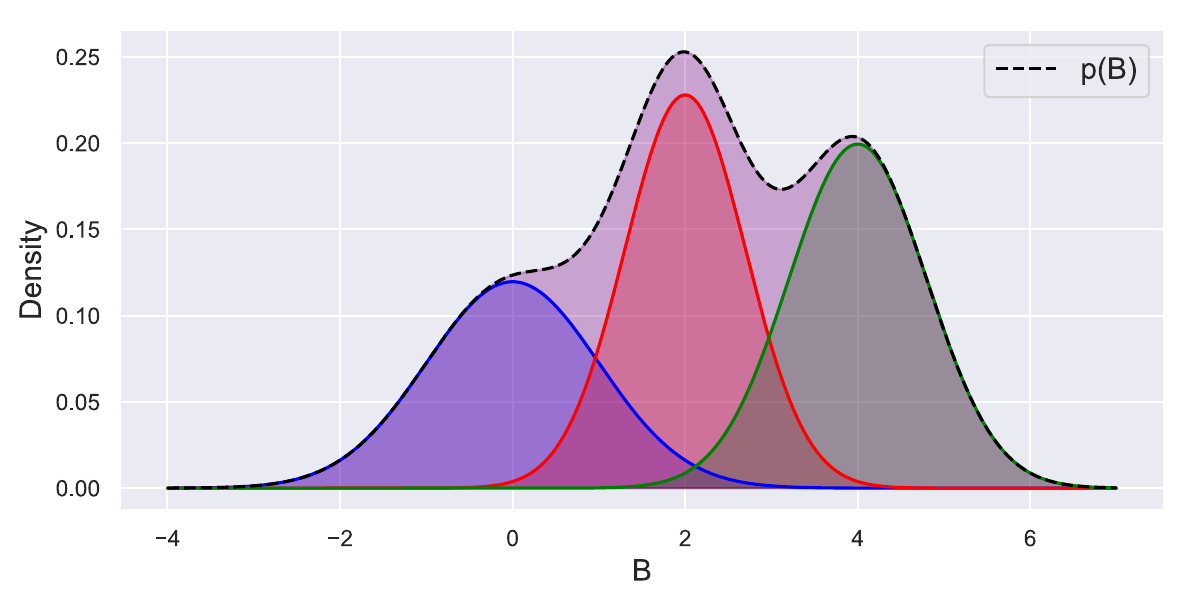}}
\caption{Example of a conditional dependency where A (categorical) determines the distribution of B. Node distributions are often not fixed a-priori, and their parameters are determined by the value of a number of categorical (parent) variables. The resulting marginal distribution can be asymmetric and multimodal.}\label{fig:second_order_uncertainty}
\includegraphics[width=\textwidth]{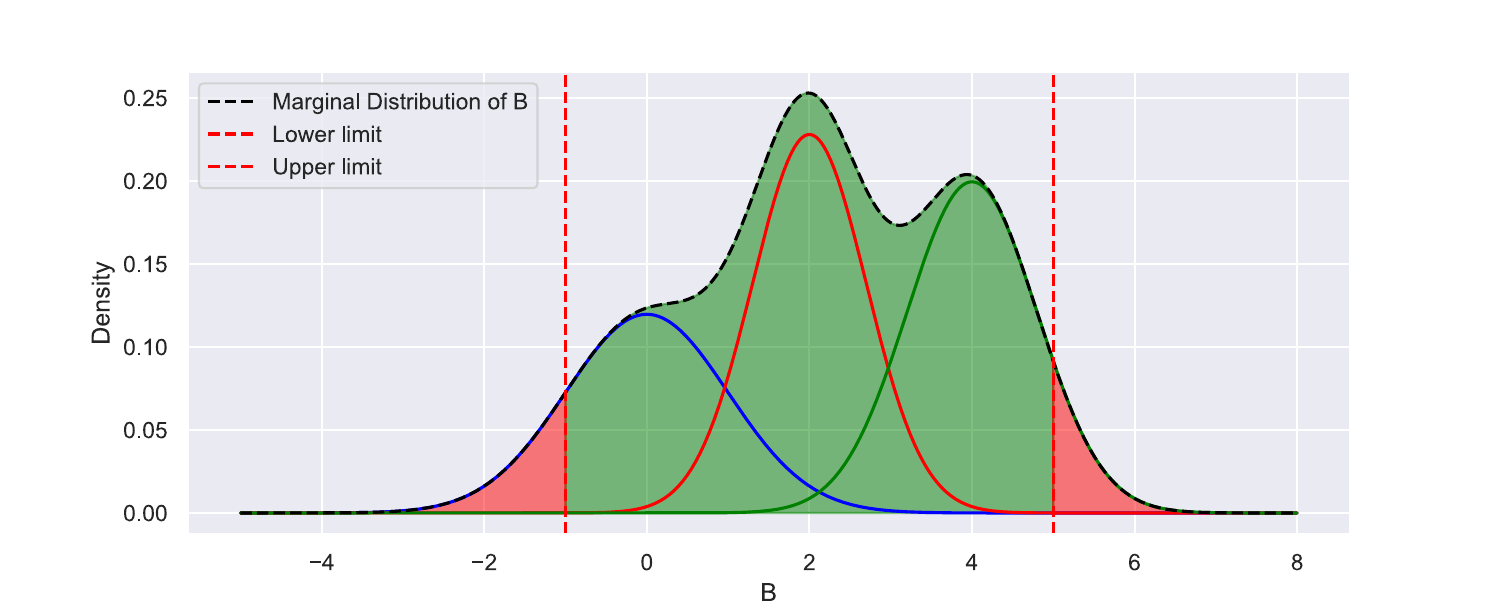}
\caption{Given a \textit{monitored} variable B, a monitoring mechanism checks if its value lies within an ideal range defined by the interval $\left[\text{B\_LTL}, \text{B\_UTL}\right]$. If yes, a binary r.v. B\_MpGood will be \textit{True}, signaling that the attribute is conformal, otherwise \textit{False}. At the end of production, all the MpGood variables are aggregated into a ProcessResult variable via a logic AND operation, which consequently signal if the whole production process did run successfully.}\label{fig:monitoring}
\end{figure}

Given a node $n_1$, its distribution may depend on the value of one or more categoricals such as the supplier or the component type. For a node $n_1$ depending on a single categorical A, we can write it mathematically as
\begin{equation}
    n_1 \sim \begin{cases}
    \mathcal{N}(\mu_0, \sigma_0) & \text{if $ A = a_0$},\\
    \mathcal{N}(\mu_1, \sigma_1) & \text{if $A = a_1$},
  \end{cases} 
\end{equation}

where $\mu_i$ and $\sigma_i$ are the mean and standard deviation of two Gaussian distributions, with $\mu_0 \neq \mu_1$ and $\sigma_0 \neq \sigma_1$. In Fig. \ref{fig:second_order_uncertainty}, we provide a graphical illustration for a simple conditional dependency.

\subsection{Structural Equations} \label{appendix:structural_equations}

Hereby we provide a more in-depth description of the production process, along with its relative physical description and structural equations. For more in-depth mathematical derivations, we address the interested reader to \cite{budynas2008shigley} and \cite{Eslami2013TheoryOE}. 

\paragraph{Model of a Magnetic Valve:}
A magnetic Valve is modeled by different parameters that describe its geometric and material properties. 
The Parameters are $E_{mv}$, describing the material elasticity of the valve, $A_{leak_{MV_{raw}}}$ describing the leakage area before starting production (A supplier may give us faulty MVs), $D_{mvMax}$ describing the maximum diameter, and $D_{mvMin}$ describing the minimum diameter, and $L_{mvPF}$ describing the axial length of the MV, coinciding with the optimal engagement length between the MV and the bore during the PF process.

All those parameters are not fixed, and are indeed randomly sampled from a distribution which conditionally depend on the type and supplier of the MV. Each combination of supplier and MV type implies a different node distribution for those parameters. This mechanism is a conditional dependency as described in \ref{sec:appendix_conditional_depepdences}. Those conditional dependencies cause the marginal node distribution of those parameters to be multimodal and asymmetric. In other words, conditional dependencies induce a mixture model on the marginal node distributions.

\paragraph{Model of an Hydraulic Unit:}
An HU is modeled with the same approach as for a MV. Indeed, an HU has the parameters $E_{hu}$ describing its elasticity and a $Force_{Lim}$ describing the force which is necessary to cause a non-zero leakage area.

On each HU we have different \textit{chambers}, and every chamber has a certain number of \textit{bores}. We model \textit{each individual} bore in the HU with a set of parameters. Specifically, we have $E_{bore}$ describing the elasticity of the bore, $D_{boreMax}$ and $D_{boreMin}$ describing its maximum and minimum diameter.
In this case, conditional dependencies appear both for the general HU parameters $E_{hu}$ and $Force_{Lim}$, but also in the parametrization of each individual bore. 

\paragraph{Intrinsic Magnetic Valve Leakage:}
A magnetic valve could be manufactured in a faulty way, resulting in \textit{intrinsic leakage} through the valve, even in the “closed state”. If quality control of the MV supplier works well, this intrinsic leakage should be zero. 
However, it may also happen that a magnetic valve gets damaged during assembly (e.g. due to high forces during press-fitting), leading to leakage through the valve itself. 
The initial intrinsic leakage of the valve as delivered by the supplier is modeled using $A_{leak_MV}$. As small intrinsic leakages are more likely than high values, and as the leakage area is continuous, we modeled a probability distribution for $A_{leak_{MV_{raw}}}$ and then used a ReLU function to cut off unrealistic negative leakage area values. 
\begin{equation}
    A_{leakMV} = \text{ReLU}(A_{leak_{MV_{raw}}})
\end{equation}

\paragraph{Total Leakage Area of a Chamber}
The total leakage area of a chamber in the Hydraulic Unit block is the sum of the leakage areas of each bore/Magnetic Valve in the chamber
\begin{equation}
    A_{leak_{tot}} = A_{leak_{Bore_1}} + A_{leak_{Bore_2}} + …,
\end{equation}
where $A_{leak_{Bore}}$ is the total leakage are per bore/Magnetic valve.

The fluid is assumed to be able to take two different leakage paths, one through the valve itself ($A_{leak_{MV}}$, see below for details) and one through the Press-Fitting connection ($A_{leak_{PF}}$).  Therefore, for a single bore, the total leakage area is the sum of the leakage though the MV and through the PF.
\begin{equation}
    A_{leak_{Bore}} = A_{leak_{MV}} + A_{leak_{PF}}
\end{equation}

\paragraph{Leakage area and geometry of the Press-Fitting Connection}

The leakage area through the Press Fitting connection $A_{leak_{PF}}$ depends mainly on the geometry of the bore and Magnetic Valve. As the cylindrical surfaces are not perfectly round, we assume an interval for the maximum $(D_{mv_{Max}}, D_{bore_{Max}})$ and minimum diameter $(D_{mv_{Min}}, D_{bore_{Min}})$, respectively.
When studying the unwanted leakage of fluid, it is important to consider the difference between minimum and maximum diameters, identified as $\Delta D$, as the may have negative consequences for the press-fitting process and result in the scrap of a product. 
\begin{equation}
\begin{split}
    \Delta D_{\text{max}} &= D_{\text{mvMax}} - D_{\text{boreMin}}, \\
    \Delta D_{\text{min}} &= D_{\text{mvMin}} - D_{\text{boreMax}}.
\end{split}
\end{equation}
To account for effects from the machine on the resulting leakage (such as acentric positioning of the valve with respect to the bore during press fitting), we introduce a machine dependent limit for resulting leakage ($LeakTolMachine$). When $\Delta D$ is higher than the threshold \textit{LeakTolMachine}, we observe a leakage (area) through the press-fitting. This phenomenon can be modeled also with a ReLU function as follows 
\begin{equation}
\begin{split}
    \Delta D_{Leak_{min}} &= \Delta D_{min} - \text{LeakTolMachine} \\
    \Delta D_{Leak_{max}} &= \Delta D_{max} - \text{LeakTolMachine}
\end{split}
\implies 
\begin{split}
    A_{Leak_{min}} &= \text{ReLU}(\Delta D_{Leak_{min}}) \\
    A_{Leak_{max}} &= \text{ReLU}(\Delta D_{Leak_{max}})
\end{split}
\end{equation}

Moreover, in real production lines, it is likely that different press-fitting machines have a different threshold for leakage due to badly adjusted press fitting processes. 
Additionally, using the coefficient $\beta_{asym}$ we can model how much the total leakage area is affected by $\Delta D_{Leak_{Min}}$ and $\Delta D_{leak_{Max}}$, respectively. 
\begin{equation} \label{eq:total_leakage_asym}
\begin{split}
    A_{\text{leakPF}} &= \beta_{\text{asym}} A_{Leak_{max}} + (1 - \beta_{\text{asym}}) A_{Leak_{min}},
\end{split}
\end{equation}
where $\beta_{asym} = 1$ means that only the maximum leakage Area $A_{leak_{MV}}$ is effective, a value of 0.5 means that minimum and maximum leakage area are weighted equally. 
\paragraph{The Press Fit process}

The PF machine applies a force which inserts the MV into a bore of the HU. Apart from inserting the MV into the bore, the force will also deform the bore. At the end of the process the bore will be deeper than before by a certain amount which is determined by the physical models (with some stochasticity). Part of the deformation is permanent, and another other part will disappear after the pressing force is removed at the end of the process, as it is related to the elasticity of the material. If the force is too high, we may cause a damage that will end in the component being scrapped.
We start by defining the effective elasticity modulus $E_{\text{eff}}$ as
\begin{equation}
    E_{\text{eff}} = \left(\frac{1}{E_{\text{bore}}} + \frac{1}{E_{\text{mv}}}\right)^{-1}
\end{equation}
where $E_{\text{bore}}$ is the elasticity of the bore and $E_{\text{mv}}$ the elasticity of the MV.
The effective elasticity is used to define the stiffness of the press-fitting machine as
\begin{equation} \label{eq:pf_stiffness}
    K_{\text{stiffPF}} = K_{\text{stiffPF}_{Ref}} \cdot \frac{\Delta D_{\text{mean}}}{K_{\text{stiffPF}_{\Delta D_{Ref}}}} \cdot \frac{E_{\text{eff}}}{K_{\text{stiffPF}_{E_{Ref}}}},
\end{equation}

where $K_{\text{stiffPF}_{Ref}}$, $K_{\text{stiffPF}_{\Delta D_{Ref}}}$, and $K_{\text{stiffPF}_{E_{Ref}}}$ are new machine-dependent parameters describing how much the reference stiffness of the PF machine $K_{\text{stiffPF}_{Ref}}$ varies linearly with $\Delta D_{\text{mean}}$ and $E_{\text{eff}}$. As before, those reference parameter are not absolute and may vary across different PF machines. Moreover, in \ref{eq:pf_stiffness} $\Delta D_{\text{mean}}$ is modeled similarly to Eq.\ref{eq:total_leakage_asym}, where we use $\beta_{asym}$ again to balance how much the PF process is affected by the maximum and minimum diameter,
\begin{equation}
    \Delta D_{\text{mean}} = \beta_{\text{asym}}\Delta D_{\text{max}} + (1 - \beta_{\text{asym}})\Delta D_{\text{min}}.
\end{equation}
Now we have all the quantities which are necessary to compute the total stiffness $K_{\text{stiff}}$ of the system,
\begin{equation}
    K_{\text{stiff}} = \left(\frac{1}{K_{\text{stiffMachine}}} + \frac{1}{K_{\text{stiffPF}}}\right)^{-1}
\end{equation}
where $K_{\text{stiffMachine}}$ is the stiffness deriving from the machine itself, and $K_{\text{stiffPF}}$ is the stiffness coming from the press-fitting operation.
Using $K_{\text{stiffPF}}$ it is possible to derive the pressing force as 
\begin{equation}
    \text{Force} = L_{\text{mvPF}} \cdot K_{\text{stiffPF}}
\end{equation}
where we used the axial length of the MV $L_{\text{mvPF}}$, as it coincides with how much the MV should be inserted into the HU with PF.
By dividing the Force by the stiffness of the system $K_{\text{Stiff}}$, we can compute the difference in vertical position of the PF tool before and after the operation, which coincides with the permanent deformation (in depth) of the component, 
\begin{equation}
    \Delta s_{\text{grad}} = \frac{\text{Force}}{K_{\text{stiff}}}.
\end{equation}
We remark that $\Delta s_{\text{grad}}$ also coincides wit the difference in position of the tool before and after the maximum pressing force is achieved and removed. Therefore, it does not include any elastic effect of the material, which may be present only while the pressing force is still present.
The quantity above can be used to compute the final position of the tool $s_{\text{grad}}$
\begin{equation}
    s_{\text{grad}} = s_0 + \Delta s_{\text{grad}}
\end{equation}
where $s_0$ is, instead, the position of the PF tool at the beginning of the process.

\paragraph{Maximum forces and dispacement on a single bore:}
As written above, during PF multiple forces are applied to insert all MVs into the HU. Focusing on the maximum force $F_{\text{max}}$ achieved on a single bore/MV pair, we can decompose it on the optimal $Force$ variable, plus another variable $\Delta F_{\text{trigger}_{\text{stop}}}$ describing how much the force went over the value Force, before a trigger in the machine did stop the operation.
\begin{equation}
    F_{\text{max}} = \text{Force} + \Delta F_{\text{trigger}_{\text{stop}}},
\end{equation}
where $\Delta F_{\text{trigger}_{\text{stop}}}$ is randomly sampled. 
The reason why we model the maximum force is because, if the applied force is too high, the component will be damaged and result in a leakage.
Moreover, from the maximum bore force we can compute the maximum difference in displacement of the tool during the PF process, written as
\begin{equation}
    \Delta s_{\text{max}} = \frac{\Delta F_{\text{trigger}_{\text{stop}}}}{K_{\text{stiffMachine}}},
\end{equation}
which, with respect to $\Delta s_{\text{grad}}$, includes also the elastic deformation which will disappear after the force is removed. Thanks to $\Delta s_{\text{max}}$ we can get the absolute maximum displacement of the tool,
\begin{equation}
    s_{\text{max}} = s_{\text{grad}} + \Delta s_{\text{max}}.
\end{equation}
The maximum displacement $s_{\text{max}}$ during the process includes both the actual deformation of the component, but also an elastic deformation which will disappear once the pressing force is removed.

\paragraph{Maximum Forces and Displacement:}
Forces applied during PF cannot be higher than a machine and product-dependent threshold $F_{\text{lim}}$, otherwise we might incur in a damage of the components. 
First, we define $F_{\text{max}}$ as the highest value achieved among all maximum forces in the chamber's bores.
Then, we can compute how much the maximum force went over the limit with
\begin{equation}
    \Delta \text{Force} = F_{\text{max}} - F_{\text{Lim}} \quad \implies \quad \Delta \text{Force}_{\text{ReLu}} = \text{ReLU}(\Delta \text{Force})
\end{equation}
where we applied a ReLU again to make it zero if the force was below the limit.
In order to model the relation between the applied forces and potential faults inducing a nonzero leakage area, we model the $\text{LeakTolMachine}$ parameter as follows:
\begin{equation}
    \text{LeakTolMachine} = \text{LeakTolMachine}_0 + \frac{\text{LeakTolMachine}_{\text{REF}} \cdot \Delta \text{Force}_{\text{ReLu}}}{\Delta \text{Force}_{\text{REF}}}
\end{equation}
where we made explicit the dependence on $\Delta \text{Force}_{\text{ReLu}}$. Lastly, we have similar machine parameters $\text{LeakTolMachine}_0$ to model the minimum tolerance, plus $\text{LeakTolMachine}_{\text{REF}}$ and $\Delta \text{Force}_{\text{REF}}$ to model the dependence on $\Delta \text{Force}_{\text{ReLu}}$.

\section{Additional Results} \label{sec:appendix_additional_results}

In this section we provide a more exhaustive exposition of our results for performance and runtime.

\begin{table*}[h] 
\centering
\begin{tabular}{ p{1.5cm}|p{1.9cm}|p{1.9cm}|p{1.9cm}|p{1.9cm}|p{1.9cm} }
 %\hline
 Causal Model & ATE MSE & CATE MSE & JS-Div Tr. & MSE & MMD\\
 \hline
 CBN   & 0.659 (0.001)  & \textbf{0.036 (0.007)} & 0.136 (0.092) & 0.259 (0.186) &  0.702 (0.121) \\
 NCM   & 0.631 (0.015)   & 0.049 (0.028) & 0.233 (0.040) & 0.307 (0.033) &  \textbf{0.086 (0.000)} \\
 CAREFL & 0.652 (0.014) & 0.175 (0.106) & 0.512 (0.093) & \textbf{0.086 (0.073)} &  Nan \\
 CNF   &\textbf{0.631 (0.015})   & 0.065 (0.063) & 0.156 (0.047) & 0.299 (0.093) &  Nan \\
 VACA  & 0.648 (0.015)   & 0.230 (0.270) & \textbf{0.033 (0.009)} & 0.059 (0.017) &   0.128 (0.000) \\
 Linear r. & 1e8 (1e10)   & - & - & -  &  -\\
 Logistic r.& 0.698 (0.066)   & - & - & -  &  -\\
 %\hline
\end{tabular}
\caption{Comparison between models for the first treatment effect estimation task on CausalMan Small with $n = 50.000$ samples and ground truth ADMG. Instabilities during sampling prevented to evaluate MMD for CNF and CAREFL, as multiple datapoints diverged to $+\infty$ as a results of training instabilities.}
\label{table:effect_estimation_small_results_2}
\end{table*}

\begin{table*}[h] 
\centering
\begin{tabular}{ p{1.5cm}|p{1.9cm}|p{1.9cm}|p{1.9cm}|p{1.9cm}|p{1.9cm} }
 %\hline
 Causal Model & ATE MSE & CATE MSE & JS-Div Tr. & MSE & MMD\\
 \hline
 NCM   & 1.115(0.118) & 1.665(0.159) & 0.206(0.005) & \textbf{0.172(0.001)} &  0.259(0.018) \\
 CAREFL & \textbf{0.982(0.223)} & \textbf{1.539(0.635)} & 0.164(0.105) & 0.279(0.197) &  Nan \\
 CNF   & 1.218(0.012) & 1.784(0.082) & 0.297(0.003) & 0.535(0.007) &  Nan \\
 VACA  & 1.214(0.009)   & 1.890(0.163) &\textbf{ 0.163(0.003)} & 0.265(0.006) &   \textbf{0.244(0.009)} \\
 Linear r. & 4.748(0.142)   & - & - & -  &  -\\
 Logistic r.& 0.992(0.015)   & - & - & -  &  -\\
 %\hline
\end{tabular}
\caption{Comparison between models for the second treatment effect estimation task on CausalMan Small with $n = 50.000$ samples and ground truth ADMG. Linear regression in this case is clearly disadvantaged due to the presence of hidden confounders and nontrivial causal mechanisms.}\label{table:effect_estimation_small_results_3_annex}
\end{table*}

\begin{table*}[h] 
\centering
\begin{tabular}{ p{1.5cm}|p{1.9cm}|p{1.9cm}|p{1.9cm}|p{1.9cm}|p{1.9cm} }
 %\hline
 Causal Model & ATE MSE & CATE MSE & JS-Div Tr. & MSE &  MMD\\
 \hline
 NCM                 & 0.580 (0.043)                  & 0.067 (0.052)                 & 0.179 (0.016)                    & 0.257 (0.017)               & 0.380 (0.008)               \\ 
CAREFL              & \textbf{0.614 (0.009)}                  & \textbf{0.033 (0.015)}                 & \textbf{0.054 (0.038)}    & \textbf{0.098 (0.069)}               & \textbf{0.212 (0.023)}               \\
CNF                 & 0.618 (0.006)                  & 0.062 (0.036)                 & 0.127 (0.056)                  & 0.218 (0.096)               & 0.335 (nan)                 \\
Linear r.  & 2e9 (2e9) &- &  -   &     -  &   -  \\
Logistic r. & 0.649 (0.119)                  & -    &      -           &    -            &           -       \\

 %\hline
\end{tabular}
\caption{Comparison between models for the first treatment effect estimation task on CausalMan Medium with $n = 20.000$ samples and ground truth ADMG. }
\label{table:effect_estimation_medium_results_1}
\end{table*}

\begin{table*}[h] 
\centering
\begin{tabular}{ p{1.5cm}|p{1.9cm}|p{1.9cm}|p{1.9cm}|p{1.9cm}|p{1.9cm} }
 %\hline
 Causal Model & ATE MSE & CATE MSE & JS-Div Tr. &  MSE &  MMD\\
 \hline
 NCM               & 1.629 (0.031)                 & 1.271 (0.031)                 & 0.589 (0.000)               & 1.000 (0.000)               & 0.389 (0.007)              \\ 
CAREFL              & 1.730 (0.068)                 & \textbf{1.199 (0.149)}                 & \textbf{0.351 (0.028)}               & \textbf{0.780 (0.034)}               & 0.185 (0.022)              \\
CNF                 & 1.822 (0.016)                 & 1.347 (0.052)                 & 0.357 (0.088)               & 0.783 (0.099)               & 0.212 (0.159)              \\
Linear r.  & \textbf{0.297 (0.019)}                 & -     &    -    &       -   &     -  \\
Logistic r. & 1.362 (0.016)                 & -   &   -   &   -    &   -   \\

 %\hline
\end{tabular}
\caption{Comparison between models for the first treatment effect estimation task on CausalMan Medium with $n = 20.000$ samples and ground truth ADMG. }
\label{table:effect_estimation_medium_results_2}
\end{table*}

\begin{table}[ht]
\centering
\begin{tabular}{ p{1.7cm}|p{2.1cm}|p{1.9cm}|p{1.9cm}|p{2.2cm}|p{2cm}} 
%\hline
Method & SHD & Prec. & Rec.& SID & p-SD\\
\hline
PC & 144.2 (0.837) & \textbf{0.123 (0.014)} & 0.056 (0.007) & 2208.2(40.935) & 0.099(0.043)\\ 
PC-Stable & 127.4 (1.949) & 0.072 (0.052) & 0.017 (0.012) & \textbf{2118.4(78.904)}& 0.017(0.004)\\
DAG-GNN & 147.8 (13.479) & 0.008 (0.017) & 0.002 (0.004) & 2275.8(32.568)& 0.038(0.017)\\
NOTEARS & 137.8 (1.922) & 0.018 (0.028) & 0.005 (0.007) & 2280.4(14.398)& 0.078(0.015)\\ 
GOLEM & 263.2 (19.791) & 0.043 (0.015) & \textbf{0.063 (0.024)}& 2371.8(40.258)& 0.427(0.003)\\ 
LiNGAM & 212.2 (31.196) & 0.043 (0.014)& 0.043 (0.022) & 2271(34.655)& 0.278(0.028)\\
GranDAG & \textbf{116 (2.646)} & 0.022 (0.049) & 0.002 (0.004) & 2240.2(24.468)& \textbf{0.001(0.001)}\\
Random DAG & 	208 (15.215) & 0.051 (0.017) & 0.050 (0.017) & 2260.8(75.652)& 0.413(0.026)\\
%\hline
\end{tabular}
\caption{Comparison for Causal Discovery on CausalMan Small (20.000 Samples).}
\label{table:causal_discovery_small}
\end{table}

\begin{table}[ht]
\centering
\begin{tabular}{ p{1.8cm}|p{2.1cm}|p{2cm}|p{2cm}|p{2.4cm}} 
%\hline
Method & SHD & Prec. & Rec. & p-SD \\
\hline
PC & 702.0 (3.24) & 0.015 (0.003) & 0.004 (0.001) & 0.061(0.05)\\ 
PC-Stable & 591.2 (0.83) & 0.020 (0.007) & 0.002 (0.001) & 0.002(0.001)\\
DAG-GNN & 	580.8 (22.28) & 0.003 (0.006) & 0.000 (0.001)& 0.001(0.001)
 \\
NOTEARS & 580.2 (1.78) & 0.024 (0.026) & 0.002 (0.002) & 0.004(0.001)\\ 
GOLEM &  845.0 (113.00) &\textbf{0.028 (0.005)} & 0.012 (0.004)& 0.283(0.131)\\ 
LiNGAM & 960.2 (100.18) & 0.027 (0.015) & 0.016 (0.007)& 0.287(0.015)\\
GranDAG & \textbf{543.4 (2.88)} & 0.017 (0.037) & 0.000 (0.001)& \textbf{2.32e-5(3.79e-5)}\\
Random DAG & 1,189.6 (9.83) & 0.020 (0.002) & \textbf{0.019 (0.002)}& 0.474(0.004)\\
%\hline
\end{tabular}
\caption{Comparison for Causal Discovery on CausalMan Medium (20.000 Samples).}
\label{table:causal_discovery_medium}
\end{table}

\begin{figure*}[ht]
\centering
\includegraphics[width=\textwidth]{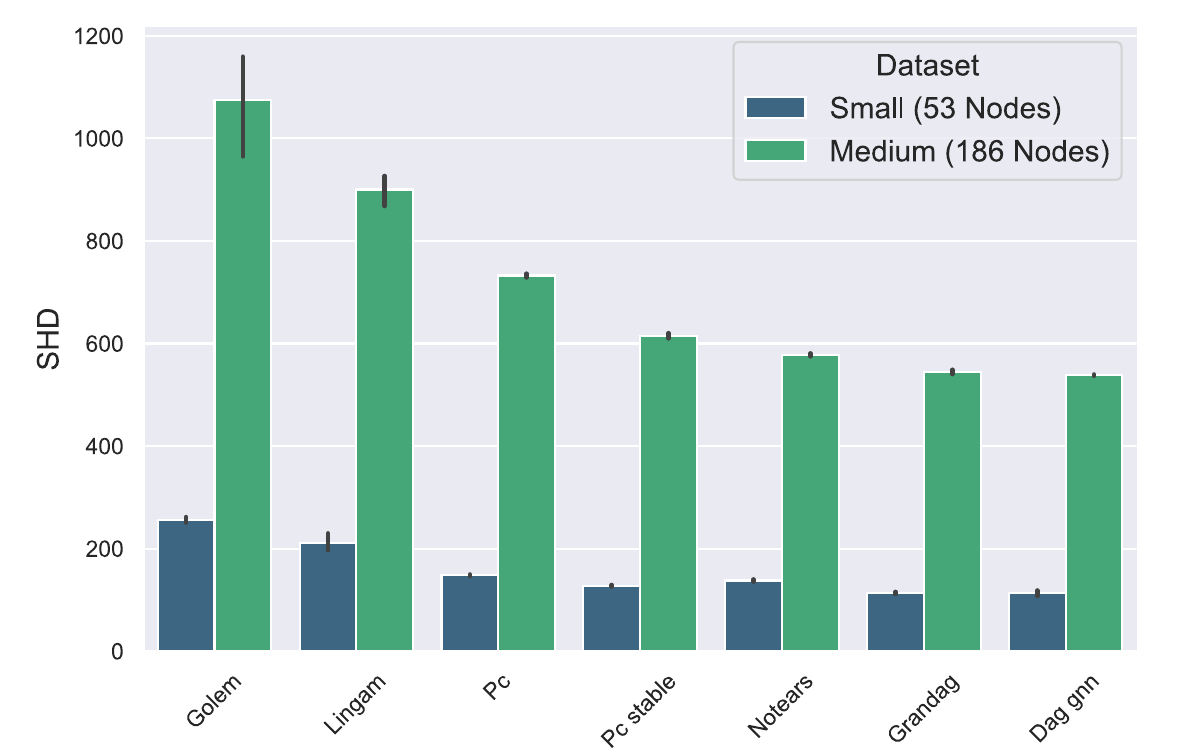} 
\caption{Difference in SHD between CausalMan Small and Medium.}
\label{fig:shd_bar_comparison}
\end{figure*}

\begin{figure*}[ht]
\centering
\subfloat[]{\includegraphics[width=0.490\textwidth]{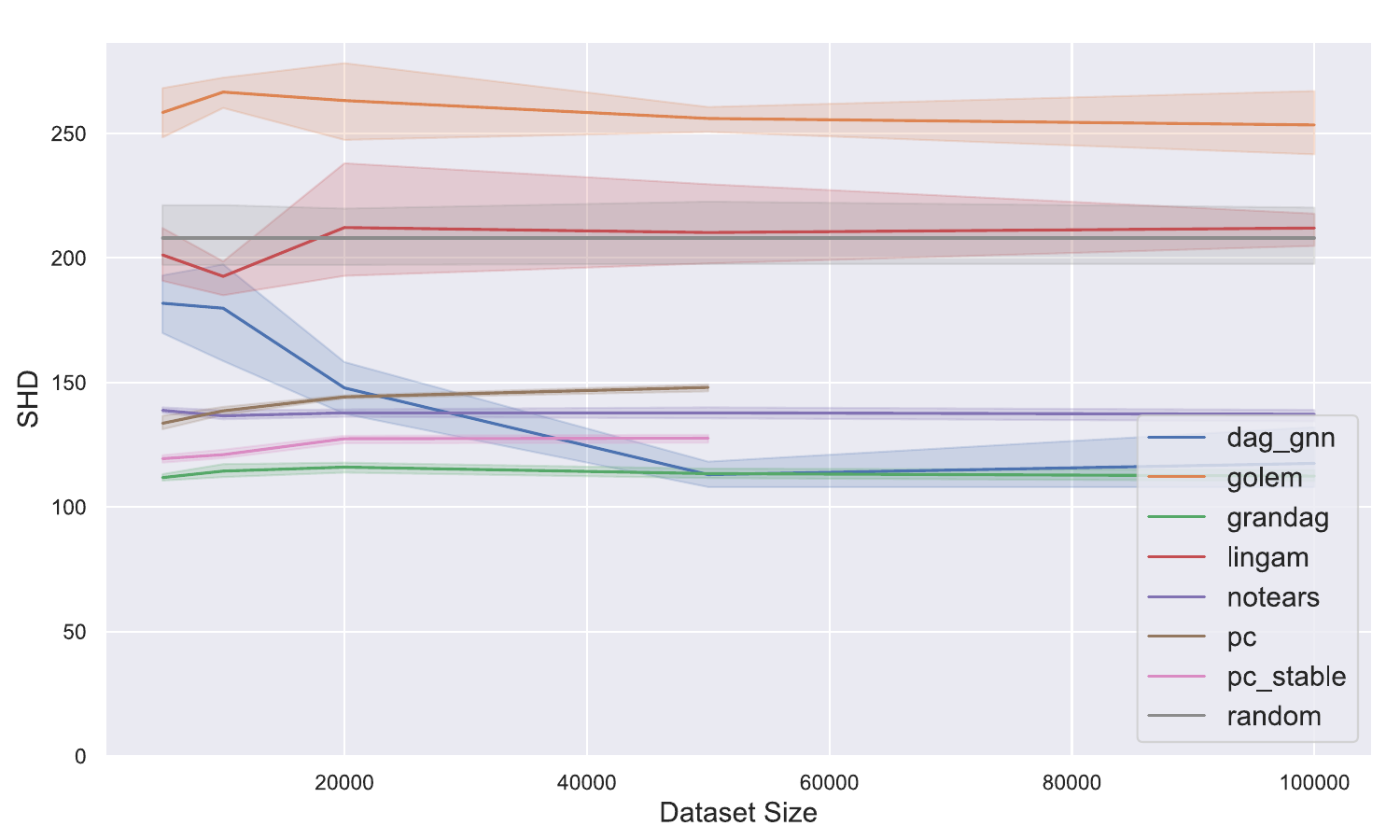} \label{fig:shd_vs_dataset_size_1}}
 \subfloat[]{\includegraphics[width=0.49\textwidth]{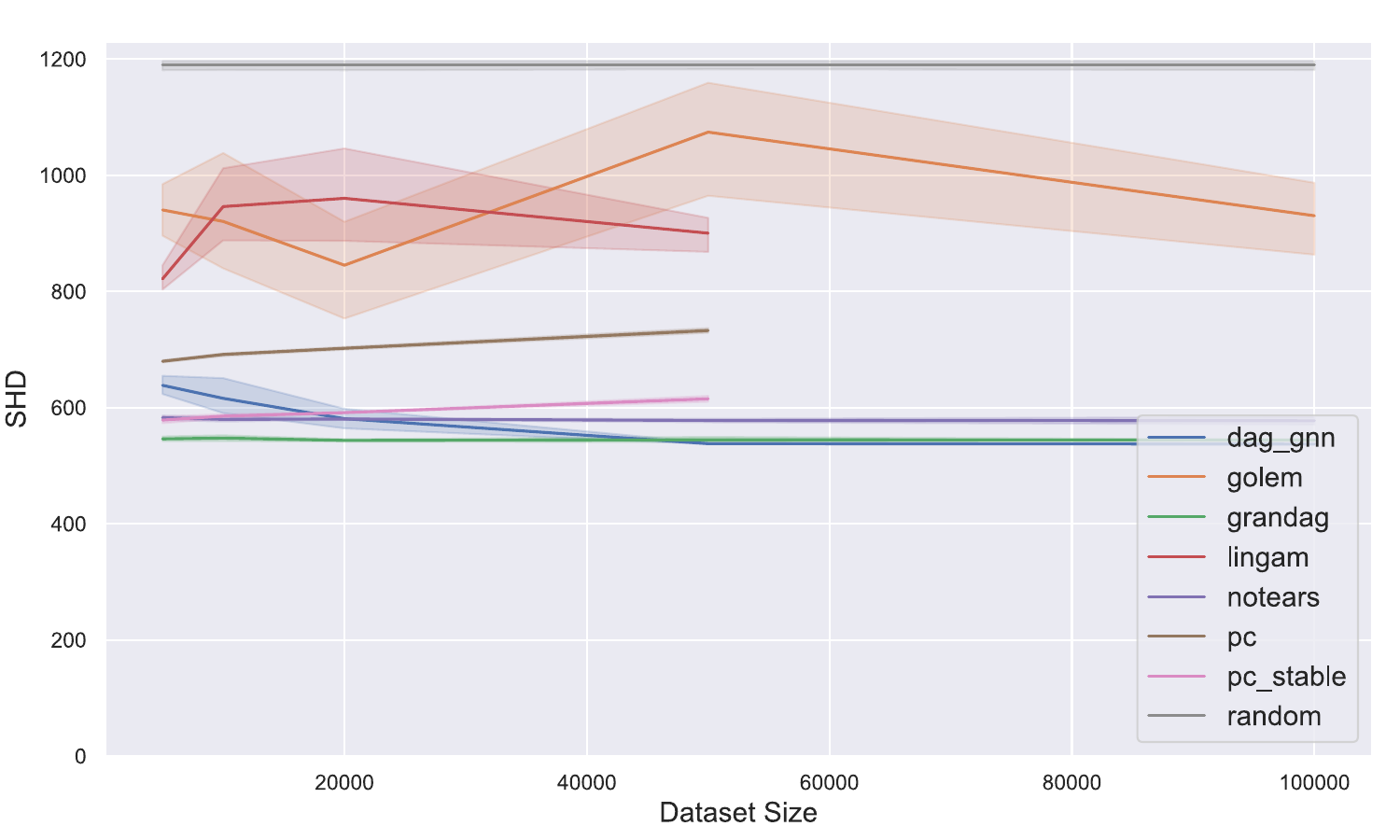} \label{fig:shd_vs_dataset_size_2}}
\caption{SHD as a function of dataset size for CausalMan Small (\ref{fig:shd_vs_dataset_size_1}) and Medium (\ref{fig:shd_vs_dataset_size_2}). Using more data has a minimal impact and is mostly detrimental to the overall Structural Hamming Distance.}
\end{figure*}

\begin{figure*}[ht]
\centering
\includegraphics[width=\textwidth]{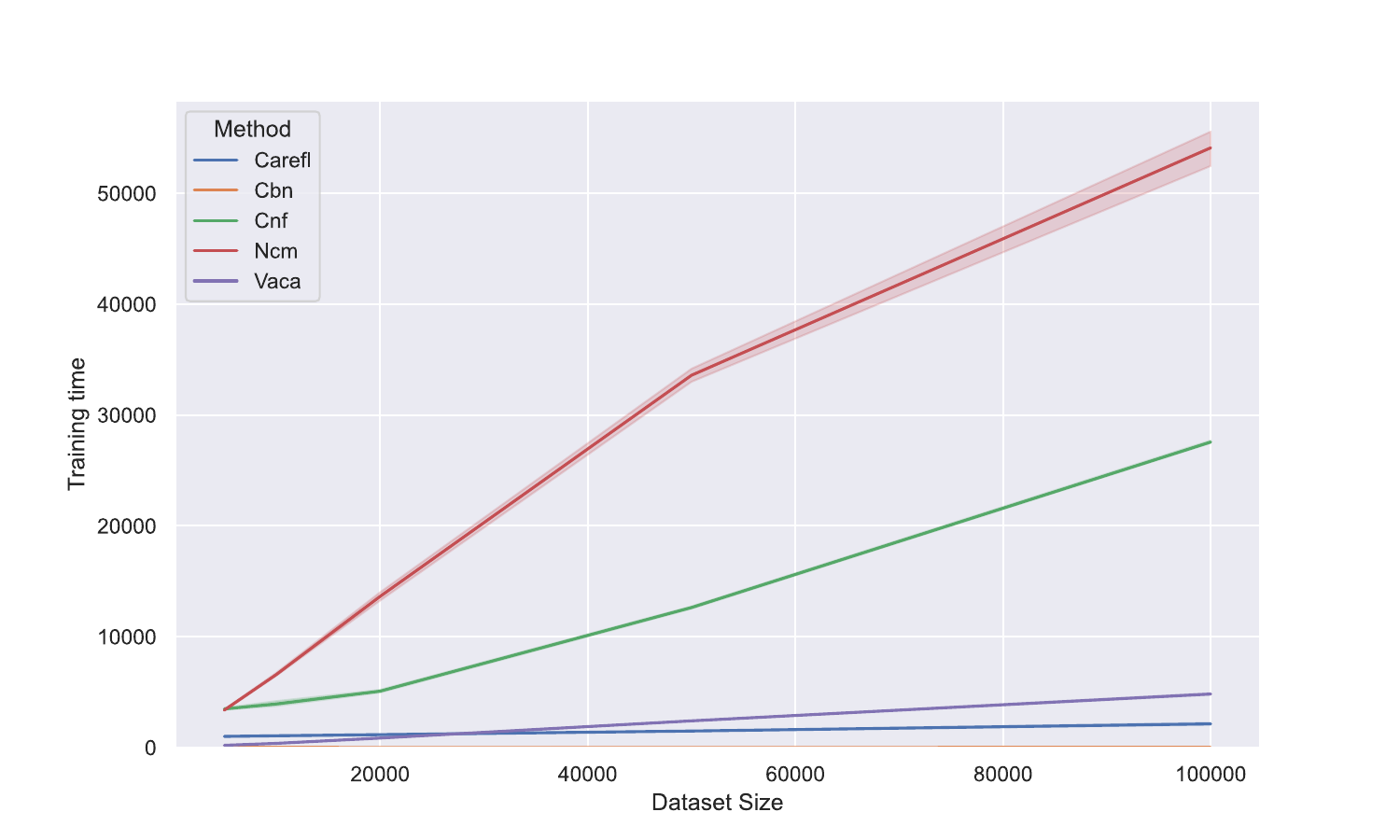} 
\caption{Average training time (seconds) vs. dataset size for CausalMan Small. }\label{fig:runtime_vs_dataset_size}
\end{figure*}

\begin{figure*}[ht]
\centering
\includegraphics[width=\textwidth]{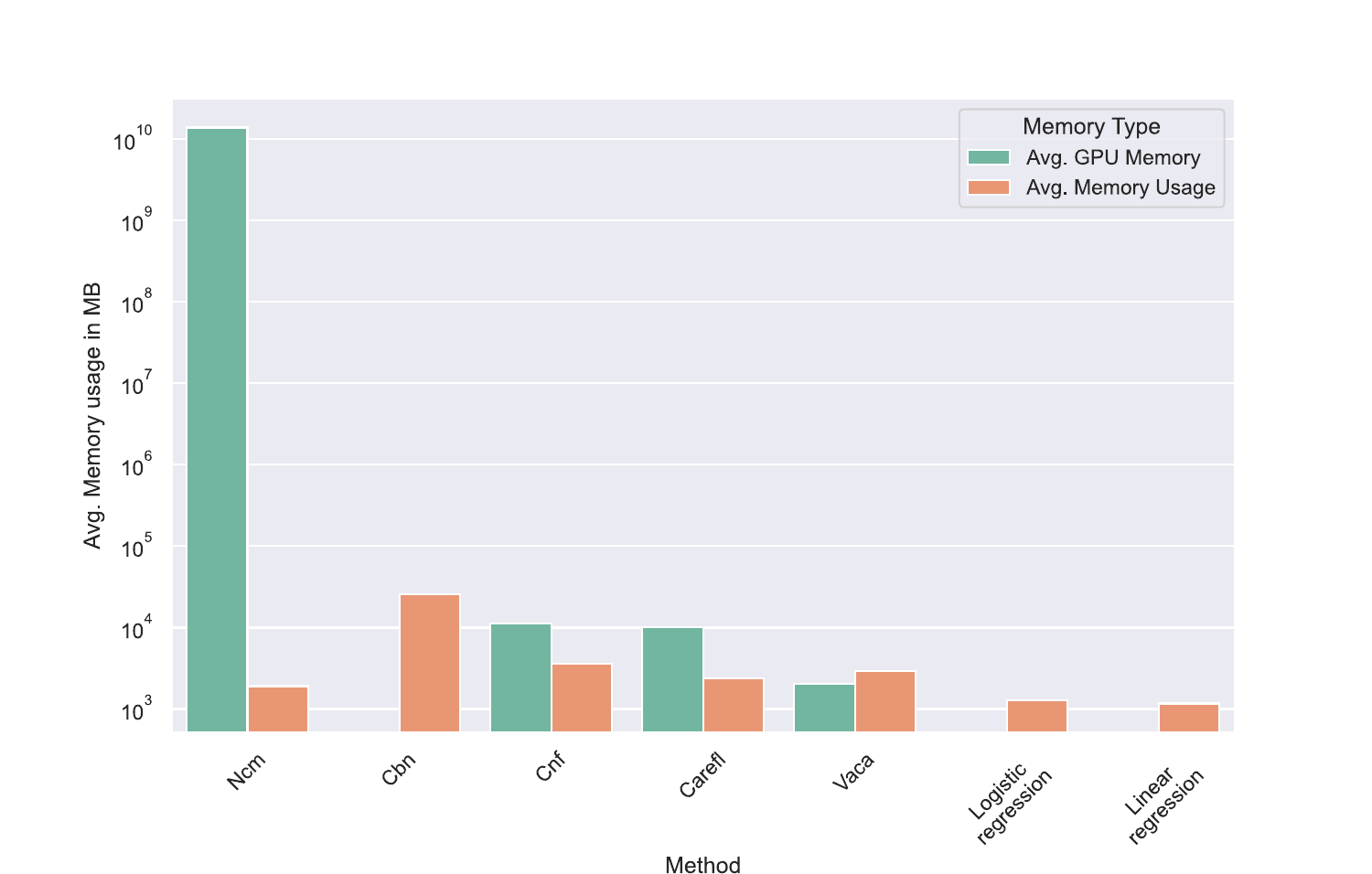} 
\caption{Bar plot showing the memory usage (RAM and GPU) for CausalMan Medium. Due to the significant time required for convergence, for NCMs it is essential to maintain a high batch size to ensure a reasonable training time. 
However, there are memory limitations when increasing the batch size, which impose a constraint on the maximum size of the dataset that NCMs can handle. 
This is a characteristic of the model that is related to the training procedure and architecture of each individual parameterized structural equation, as shown in~\cite{notallcausalinferencezece}. For example, a sum-product network is $\mathcal{O}(n)$ for a forward pass with $n$ the number of parameters, whereas MADE \cite{pmlr-v37-germain15} is approximately $\mathcal{O}(n^2)$.}
\label{fig:memory_usage_dataset1}
\end{figure*}

\section{Experiment Setting}

\subsection{Metrics} \label{sec:appendix_metrics}

We can write the \textit{Structural Hamming Distance} SHD between a graph $\displaystyle \gG$ with adjacency matrix $A$ and the ground truth $\displaystyle \gG^\star$ with adjacency matrix $A^\star$ as:
\begin{equation}
    SHD(A, A^\star) = \sum_{i,j = 0}^n \mathbf{I}_{A_{ij} \neq A^\star_{ij}}
\end{equation}

Since discovering an individual edge can be thought as a binary classification task (edge/no-edge), it  is common to measure metrics such as precision and recall:

\begin{align}
    Pr &= \frac{tp}{tp + fp}, &
    Rec &= \frac{tp}{tp + fn}.
\end{align}

where \textit{tp} stands for true positives, \textit{fp} for false positives and \textit{fn} for false negatives.

\subsection{Causal Models} \label{sec:appendix_causal_models}

Here we provide a more detailed description of the tested causal models.

\begin{itemize}
    \item \textbf{Causal Bayesian Networks:} For Bayesian Networks (BN), edges do not have a causal semantic, and they are indeed only an observational Layer 1 model. However, it is possible to define a do-operator for Bayesian Networks, and obtain an interventional L2 model called \textit{Causal Bayesian Network} (CBN) \citep{bareinboimpch}.
    \item \textbf{Neural Causal Models:} Presented by \cite{xia2022causalneuralconnectionexpressivenesslearnability}, \textit{Neural Causal Models} (NCM) consist in a SCM where each structural equation is parameterized by a neural network. NCMs, as a special case of SCMs, are Layer 3 models capable of answering counterfactual queries, when identifiable \citep{xia2022neuralcausalmodelscounterfactual}. More info about our implementation in \ref{sec:appendix_code}.
    \item \textbf{CAREFL:} \textit{Causal AutoREgressive normalizing Flows} (CAREFL) \citep{careflkhemakhem2021}, uses Normalizing flows with affine layers and the Causal Ordering to answer queries up to the counterfactual level.
    \item \textbf{Causal Normalizing Flows:} \citep{causalnormjavaloy2023} provided a generalisation of CAREFL that uses the whole causal graph, includes non-additive noise models, and provides stronger identification guarantees, yielding Causal Normalizing Flows (CNF).
    \item \textbf{VACA:} Based on \textit{Variational Graph Autoencoders} \citep{kipf2016variationalgraphautoencoders}, \textit{Variational Causal Graph Autoencoder} (VACA) \citep{vacasanchezmartin2021} provides a counterfactual model based on \textit{Graph Neural Networks}.
\end{itemize}

\section{Implementation details}\label{sec:appendix_implementation}

In this supplementary section, we provide additional details on the architectures and implementations that have been tested. Furthermore, we list all the necessary modification that have been necessary to run the models with our datasets with hybrid data-types.

\subsection{Determinism}

Every experiment was run 5 different times with the random seeds 4, 6, 42, 66 and 90. 

\subsection{Hardware}
To perform a fair experimental evaluation of their tractability, each run was performed on a A100 GPU with 80 GB of GPU memory allocated, and an AMD EPYC 7643 CPU, with approximately 300 GB of RAM memory allocated.

Not all methods can leverage GPU parallelisation, therefore:

\begin{itemize}
    \item For Causal Inference, regression-based techniques and CBNs are run using only CPUs. 
    \item For Causal Discovery, PC algorithm, PC-Stable, NOTEARS, and LiNGAM are run using only CPUs.
\end{itemize}

\subsection{Data preprocessing:} For running the chosen models, data had to be embedded in a numerical format. Therefore, categoricals and discrete variables have been converted to an ordinal encoding (1, 2, 3, etc.). After obtaining a purely numerical dataset, every individual variable has been normalized via min-max normalization to be within the -1 and 1 range. 

Models like CBNs are designed to work exclusively on discrete domain and are not tailored for hybrid datatypes.
To overcome this limitation, CBNs have been fitted on a different version of the datasets where the continuous variables have been uniformly quantized.

For CNFs, CAREFL and VACA, data sampled from those models had to receive a binarization of the outcome variable and a binning of the conditioning variable. The binarization of the target variables has been done such that the target variable would be -1 if output was less than 0, and 1 if output higher than 0. For the conditioning variable, instead, the bins corresponded to the values of the evidence variable that were present in the training data, and the operation was necessary since the variable is discrete, otherwise it would have been impossible to evaluate empirically the conditional interventional distribution.

Among our tested models, only NCMs can adapt by design to hybrid datatypes, therefore they are the only ones that didn't necessitate any pre-processing for the training data apart from embedding of categoricals and data normalization. During estimation, interventional distributions were computed directly from the raw data that has been sampled from the estimated interventional distributions, without any post-processing.

\subsection{Implementation of Causal Models} \label{sec:appendix_code}

For convenience, all the tested models have been incorporated into a configurable framework, present in this paper's supplementary material.

\paragraph{Linear and Logistic Regression:} For linear and logistic regression estimates, we used the implementations provided in the \textit{DoWhy} python library. 

\paragraph{Causal Bayesian Networks (CNB):} For CBNs, we use the implementation contained in the \textit{pgmpy} python library. We use the K2 score function.

\paragraph{Neural Causal Models (NCM):} We used the original implementation contained in \href{https://github.com/CausalAILab/NeuralCausalModels}{Github Link}, and applied minor modifications in order to adapt the model to handle hybrid data-types. Modifications have been made because, for each individual parameterized structural equation, NCMs require architectures capable of estimating conditional distributions $p(v_i|\displaystyle \parents_\gG(v_i), u_i)$, as their log-likelihood is used for training \cite{xia2022causalneuralconnectionexpressivenesslearnability}.
In detail, binary variables have been modeled using MADE, as in the original paper. The MADE implementation we use is taken from: \href{https://github.com/karpathy/pytorch-made}{Github Link}.
For discrete/categorical variables, MADE is still used upon minor modifications to the architecture in order to adapt it to discrete and non-binary domains. Indeed, discrete variables have been one-hot-encoded, then fed to the neural network, which would output the logit values for each discrete value. The input size of MADE in this case would be, for a causal graph $\displaystyle \gG$, 
\begin{equation}
    D = |\displaystyle \parents_\gG(\ervx_i)| + |u_i| + |v_i|.
\end{equation}

where the last $|v_i|$ variables consist in the one-hot-encoding of the realisation of $v_i$.

Finally, structural Equations for Continuous variables are parameterised using Conditional Normalizing Flows
\cite{winkler2023learninglikelihoodsconditionalnormalizing}

\paragraph{Causal Normalizing Flows, CAREFL \& VACA:} We use the original author's implementation that can be found at \href{https://github.com/psanch21/causal-flows}{Link to GitHub Repository}.

\subsection{Hyper-parameters and Training Settings}

To ensure reproducibility of every experiment, we list here all the modification applied to every single causal model and causal discovery method.

\subsubsection{Settings for Causal Models}

We reflect the implementation used in the original papers for all Causal Models tested. However, given the large size of the dataset in terms of covariates and number of datapoints, we apply the following modifications, mostly to increase the number of parameters and capacity for each model. Modifications are as follows:
 
\begin{itemize}
    \item \textbf{CAREFL and Causal Normalizing Flows}: For both models, we did increase their size to have 4 layers with 64 hidden nodes each. Training optimization parameters are not changed with respect to the paper \citep{causalnormjavaloy2023}.
    \item \textbf{VACA}: 300 training epochs and batch-size of 1024. Both encoder and decoder use the \textit{Graph Isomorphism Network} (GIN) \citep{xu2018how} version of VACA. The encoder uses 2 hidden layers. The inner dimensionality is always 64. Even tough design conditions require to have a number of layers proportional to the diameter of the graph, scaling attempts to make the model bigger resulted in loss of convergence during training, and are a common limitation of Graph Neural Networks.
    \item \textbf{NCM}: For CausalMan Small, we use a batch-size of 1024 and 1.000 training epochs. For CausalMan Medium, we use a batch-size of 2048 and 600 training epochs. Training algorithm is still \textit{AdamW} \citep{loshchilov2019decoupledweightdecayregularization} with learning rate 0.004 and the \textit{Cosine Annealing} scheduler with warm restarts.
\end{itemize}
 
\subsection{Settings for Causal Discovery} \label{sec:appendix_code_discovery}
 
All the tested models used the implementations present in the gcastle python library. All used Causal Discovery models reflect their original papers cited in \ref{sec:causal_discovery} apart from the design choices listed below:
\begin{itemize}
    \item \textbf{PC and PC-Stable:} The $\chi^2$ Conditional Independence test was used.
    \item \textbf{NOTEARS:} The $L_2$ loss function was used.
    \item \textbf{GranDAG:} We used a batch-size of 1024 samples and 4 hidden layers, each one with 64 hidden nodes.
    \item \textbf{DAG-GNN:} We used a batch-size of 1024.
\end{itemize}

\section{Ground Truth Causal Graphs}

In this section we provide a visual depiction of all the ground truth causal graphs, both the complete graphs involved in the DGP and the partially observable ones obtained after a latent projection.

\begin{figure}[h]
\centering
\includegraphics[width=\textwidth]{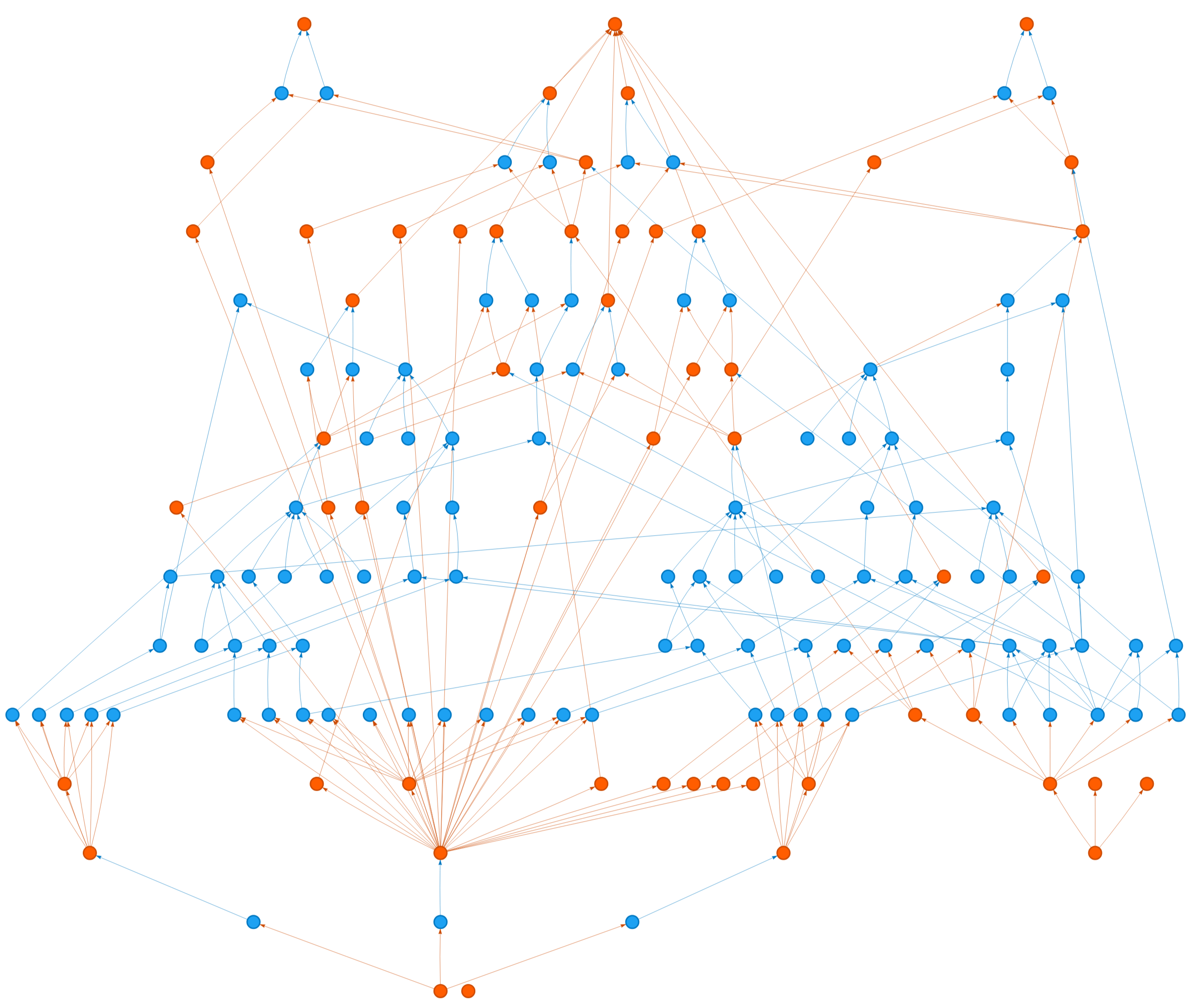}\label{fig:ground_truth_dataset0_annex}
\caption{Complete Ground truth causal graph including hidden variables for CausalMan Small. Observable variables are colored in orange, and latent ones are colored in blue. 104 of 157 (66.2\%( of variables are latent.}
\end{figure}

\begin{figure}[ht]
\centering
\includegraphics[width=0.6\textwidth]{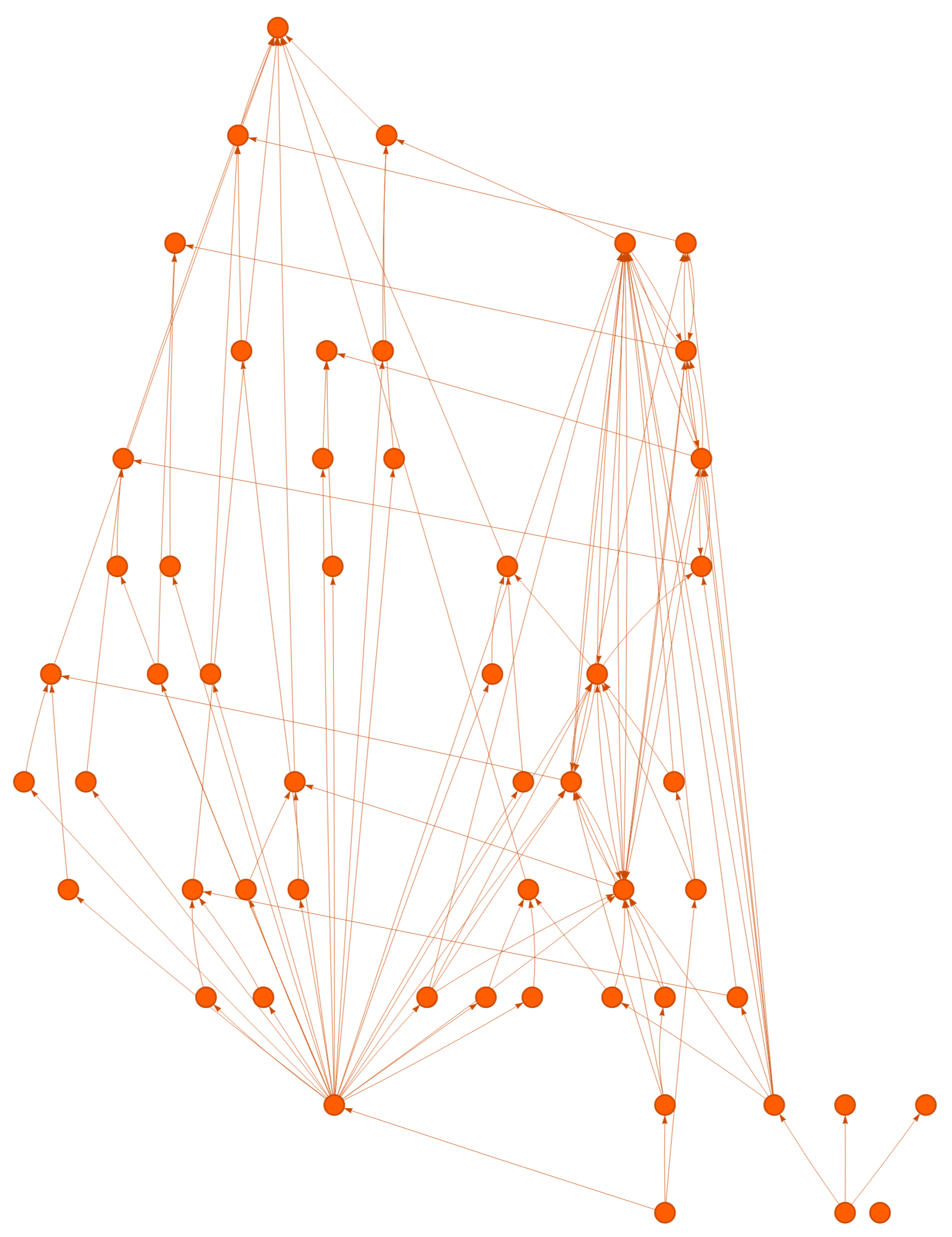}\label{fig:ground_truth_partially_observable_dataset0_annex}
\caption{Partially observable Ground truth causal graph for CausalMan Small.}
\end{figure}

\newpage
\begin{figure}[ht]
\centering
\includegraphics[angle=90, width=0.7\textwidth]{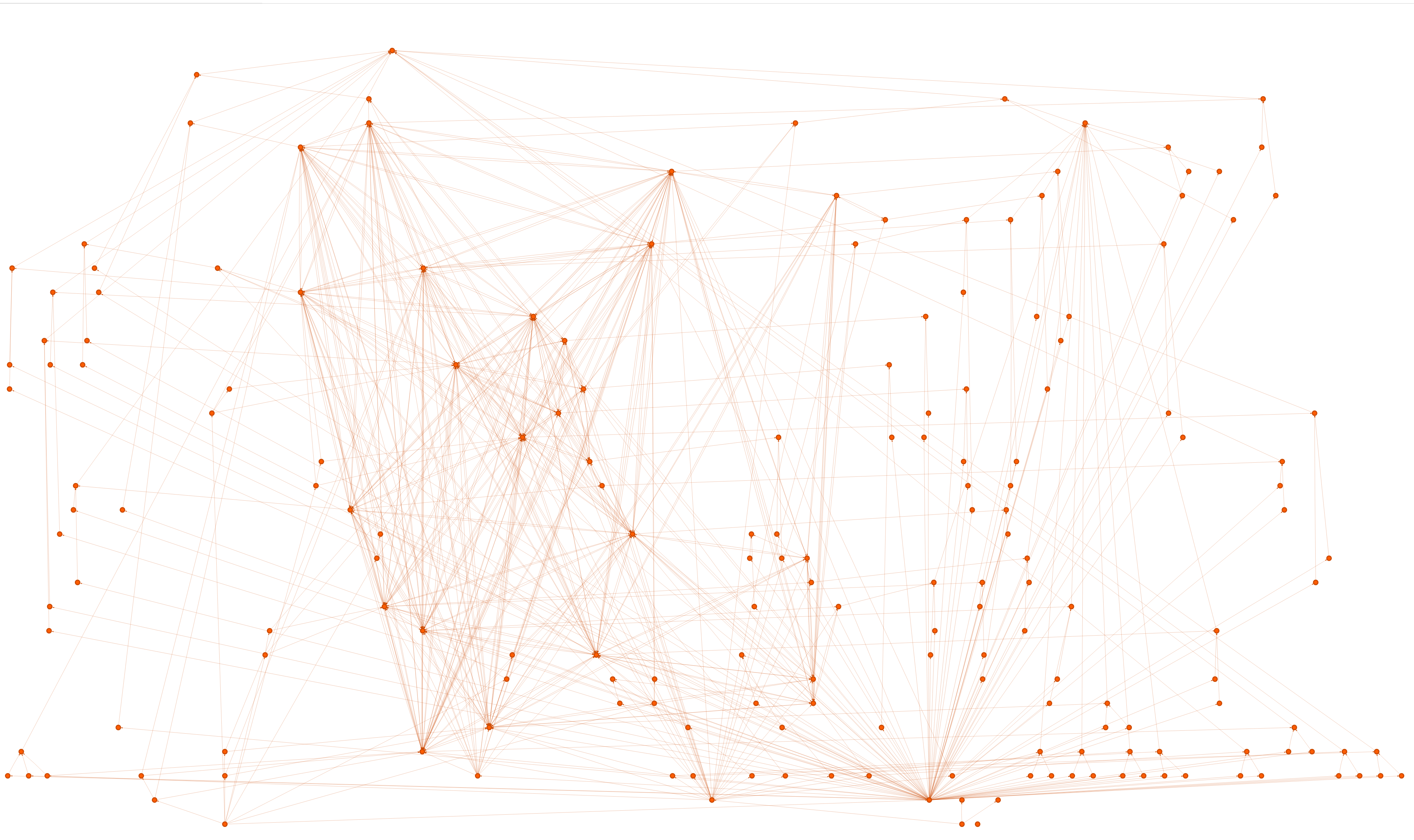}\label{fig:ground_truth_partially_observable_dataset1_annex}
\caption{Partially observable Ground truth causal graph for CausalMan Medium.}
\end{figure}

\newpage
\begin{figure}[ht]
\centering
\includegraphics[angle=90 , width=0.475\textwidth]{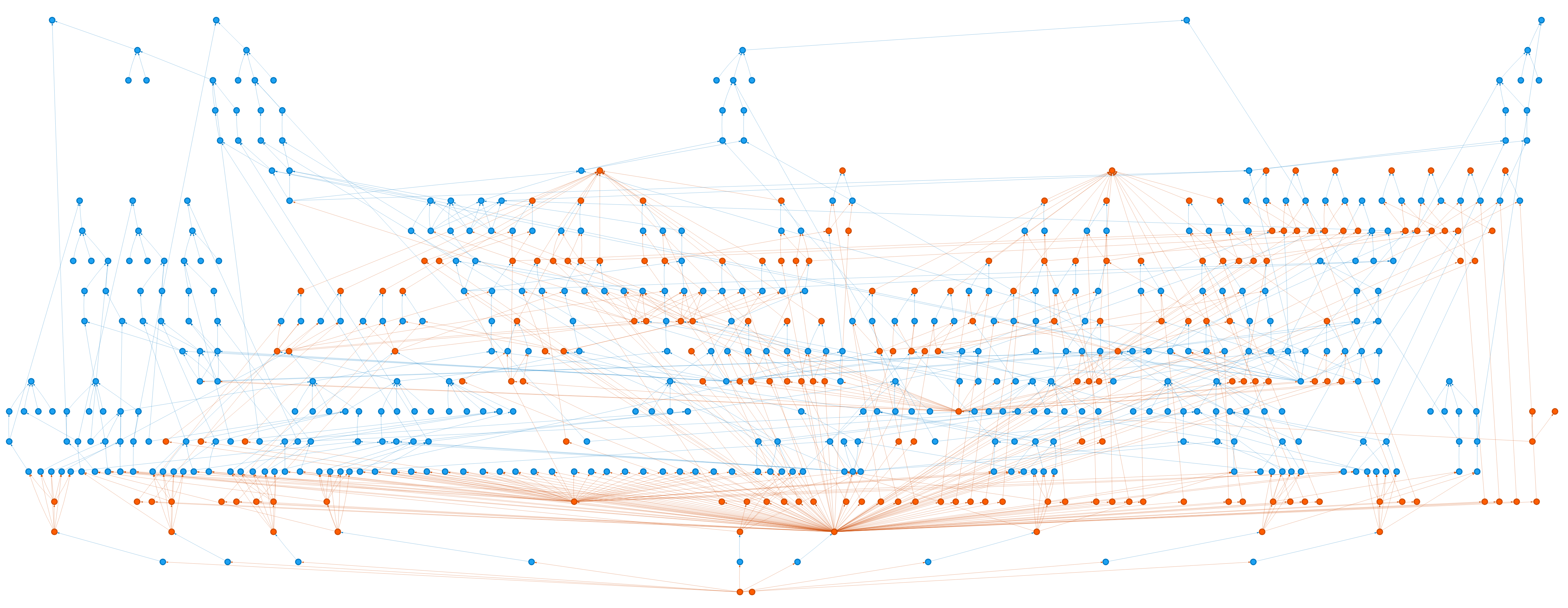}\label{fig:ground_truth_dataset1_annex}
\caption{Complete Ground truth causal graph including hidden variables for CausalMan Medium. Observable variables are colored in orange, and latent ones are colored in blue. 419 of 605 (69.2\%) of variables are latent.}
\end{figure}
%%%%%%%%%%%%%%%%%%%%%%%%%%%%%%%%%%%%%%%%%%%%%%%%%%%%%%%%%%%%%%%%%%%%%%%%%%%%%%%
%%%%%%%%%%%%%%%%%%%%%%%%%%%%%%%%%%%%%%%%%%%%%%%%%%%%%%%%%%%%%%%%%%%%%%%%%%%%%%%

\end{document}